\documentclass{article}

\usepackage{arxiv}

\usepackage[utf8]{inputenc} 
\usepackage[T1]{fontenc}    
\usepackage{amsfonts}       
\usepackage{amsmath}       

\usepackage{nicefrac}       
\usepackage{doi}

\usepackage{tikz} 
\usepackage{mathtools}
\usepackage{caption}
\usepackage{setspace}
\usepackage{bbm}
\usepackage{pdflscape}
\usepackage{adjustbox}
\usepackage{nicefrac}  
\usepackage{microtype}  
\usepackage{efbox,graphicx}
\usepackage{float}
\usepackage{pdfpages}
\usepackage[ruled,vlined]{algorithm2e}
\def\equationautorefname~#1\null{(#1)\null}

\usepackage{amsthm}
\usepackage{hyperref}
\usepackage{cleveref}

\crefname{lemma}{Lemma}{Lemmas}
\crefname{definition}{Definition}{Definitions}
\crefname{proposition}{Proposition}{Propositions}
\usepackage{enumitem}
\theoremstyle{definition}

\newlist{enumthm}{enumerate}{1}
\setlist[enumthm]{label=(\alph*)}
\newcommand{\defeq}{\vcentcolon=}
\newcommand{\eqdef}{=\vcentcolon}

\usepackage{sectsty}

\usepackage[style=authoryear-comp,
maxbibnames=10
]{biblatex}
\renewbibmacro{in:}{}
\DeclareFieldFormat
  [article,inbook,incollection,inproceedings,patent,thesis,unpublished]
  {title}{#1\isdot}

\DeclareFieldFormat{pages}{#1}

\renewbibmacro*{volume+number+eid}{%
  \printfield{volume}%
  \setunit*{\addnbspace}
  \printfield{number}%
  \setunit{\addcomma\space}%
  \printfield{eid}}

\DeclareFieldFormat[article]{number}{\mkbibparens{#1}}

\AtEveryBibitem{%
  \clearfield{issn} 
  \clearfield{doi} 

  \ifentrytype{online}{}{
    \clearfield{url}
  }
    \ifentrytype{online}{}{
    \clearfield{urlyear}
  }
}

\title{Text analysis and deep learning: A network approach}

\date{November 9, 2021, this version \today}

\author{{Ingo Marquart}\thanks{Corresponding Author} \\
	statworx GmbH\\
	\texttt{ingo.marquart@statworkx.com} \\
	\And
	{Nghi Truong}\\
	ESMT Berlin\\
	\texttt{nghi.truong@esmt.org} \\
	\And
	{Wonjae Lee}\\
	Korea Advanced Institute of Science and Technology\\
	\texttt{ingo.marquart@statworkx.com} \\
	\And
	{Matthew Bothner}\\
	ESMT Berlin\\
	\texttt{matthew.bothner@esmt.org} \\
	\And
}

\renewcommand{\shorttitle}{Text Analysis and Deep Learning}

\hypersetup{
pdftitle={Text analysis and deep learning: A network approach},
pdfsubject={},
pdfauthor={Ingo Marquart, Nghi Truong, Wonjae Lee, Matthew Bothner},
pdfkeywords={Semantic Networks, Deep Learning, Transformers, Lexical Semantics},
}

\begin{document}
\maketitle

\begin{abstract}
	Much information available to applied researchers is contained within written language or spoken text. Deep language models such as BERT have achieved unprecedented success in many applications of computational linguistics. However, much less is known about how these models can be used to analyze existing text. We propose a novel method that combines transformer models with network analysis to form a self-referential representation of language use within a corpus of interest. Our approach produces linguistic relations strongly consistent with the underlying model as well as mathematically well-defined operations on them, while reducing the amount of discretionary choices of representation and distance measures. It represents, to the best of our knowledge, the first unsupervised method to extract semantic networks directly from deep language models. We illustrate our approach in a semantic analysis of the term "founder". Using the entire corpus of Harvard Business Review from 1980 to 2020, we find that ties in our network track the semantics of discourse over time, and across contexts, identifying and relating clusters of semantic and syntactic relations. Finally, we discuss how this method can also complement and inform analyses of the behavior of deep learning models. \end{abstract}

\section{Introduction}
Recent advances in the application of deep neural networks to natural language processing have led to record-setting performance increases in practically all generative and supervised tasks, including language generation, translation, and question-answering  (\cite{brownLanguageModelsAre2020,lewisBARTDenoisingSequencetoSequence2019,liuVeryDeepTransformers2020,zaheerBigBirdTransformers2021}).
Much less is known about how these models can be used to infer knowledge from existing texts. Given that much of the information available to researchers resides in text, the objective to which we seek to contribute with the current work is the development of unsupervised methods employing these deep language models. Here, we focus on the analysis of semantics: the meanings that are conveyed and developed through natural language, and that represent the heart of the corpus.

The novel transformer architectures at the base of virtually all recent successes in computational linguistics, draw upon much more sophisticated representations of text than do earlier models  (\cite{tenney_bert_2019}).  In particular, they are able to account for the contexts in which words appear. For instance, the model’s representation of the word “founder” would differ whether the person being discussed is the creator of a company, the head of a hierarchy, or the instigator of a political movement. Such contextual awareness holds great promise for the analysis of the meaning of words in a text: its lexical semantics. Indeed, these advances in computational modeling mirror the developments in linguistics, where a strict division between lexical- and compositional semantics was eventually considered too rigid, and where contextual polysemy became a core issue (\cite{goddard2010semantic}).

However, the contextual representations of transformer models are distributed across model layers in a non-isotonic and non-Euclidian fashion, which is also corpus-, training- and instance-dependent (\cite{rogersPrimerBERTologyWhat2020}). Consequently, there is no obviously correct, generally accepted approach to working with word representations of deep language models. Despite the theoretical superiority of transformer models, their applications in semantic analyses have, thus far, failed to beat prior models (\cite{schlechtwegSemEval2020TaskUnsupervised2020}).

In this paper, we develop an unsupervised implementation that combines transformer models with the tools and methods from network theory. In particular, we build on the idea of semantic substitutions, that has been used by prior works (e.g.,\cite{alagicRepresentingWordMeaning2021,giulianelliAnalysingLexicalSemantic2020})  to analyze word senses of single tokens. We use a formal approach to analyze these implied relations between words conditional on context and develop a network representation of substitution semantics. These networks capture the self-referential information that the model provides in the absence of exogenous labels or categorizations.
To the best of our knowledge, our approach is the first to recover the contextual semantic similarity and distance relationships between words from transformer models.

As a demonstration, we analyze the roles, contexts and social identities associated with the word “founder” in the entire corpus of \textit{Harvard Business Review} from 1980 to 2020.

\subsection{Related work and contribution}

The meaning of a word is a fundamental object of study in computational linguistics, where researchers have in particular focused on the disambiguation and identification of lexical semantic change (e.g., \cite{tahmasebiSurveyComputationalApproaches2019, schlechtwegWindChangeDetecting2019}) and word senses (e.g., \cite{amramiBetterSubstitutionbasedWord2019,baskayaAIKUUsingSubstitute2013,dailleAmbiguityDiagnosisTerms2016}). Similarly, changes in meaning structures have been examined in the field of computational sociology (e.g., \cite{ruleLexicalShiftsSubstantive2015,kozlowskiGeometryCultureAnalyzing2020,linzhuoSocialCentralizationSemantic2020,padgettPoliticalDiscussionDebate2020,schootsConflictRevoltName2020}), although these latter studies are not based on deep language models and therefore do not take context or polysemy into account.

The present work aims to provide a relational characterization of language use specific to an existing corpus. The contributions of our approach can be summarized as follows:

First, we develop a formal model of substitute relationships extracted from BERT. As a formal model, our approach implies distance measures and aggregation operations that are well-defined and require no parameterization, discrete subsets of substitutes, or dimensionality reduction techniques. Deriving semantic relations via lexical substitutions (\cite{alagicRepresentingWordMeaning2021,mccarthyEnglishLexicalSubstitution2009}) has crucial advantages over vector space representations (e.g., \cite{giulianelliAnalysingLexicalSemantic2020,liuStatisticallySignificantDetection2021}).

Second, our approach goes beyond \cite{arefyevBOSSemEval2020Task2020} and \cite{amramiBetterSubstitutionbasedWord2019}, who also employ lexical substitutions. We derive semantic relations between words, rather than between occurrences of selected tokens. That is, we extract contextual semantic networks from BERT and identify semantic structures not only across the senses of a single focal word, but also  across distinct tokens. For example, we identify a syntactic-semantic cluster of \textit{roles}, in which hierarchies of semantically-related roles are embedded. Focusing on tokens substituting for alters, rather than on substitutions of occurring tokens, increases recall and allows the analysis of words that appear only infrequently or not at all in the context of interest.

Third, lexical substitution semantics are recursive. Our contextual network representations, and associated tools and methods from network analysis, characterize this higher-order interdependence. This includes changes in a word’s meaning when a focal token’s direct substitutes remain constant, but their meaning changes. For example, a stable association between “founder” and “leader” may still represent semantic shifts, if the meaning of “leadership” changes from an authoritarian to a collaborative concept. In turn, network techniques also allow us to understand global properties of language in the corpus. For example, we derive measures such as semantic importance, semantic generality, and conventionality from the structure of the relations we extract.

Fourth, computational linguistics typically examines semantic change between predefined domains and contexts, for example, whether a word gains meanings during a given interval  (\cite{tahmasebiSurveyComputationalApproaches2019}). Instead, our approach infers such semantic contexts from the text. That is, our method allows us to analyze relationships between contexts, or between contexts and focal words jointly. We can identify contexts that share a consistent semantic structure of language, be it in a single corpus or across multiple times and domains.

In summary, the present model is, to the best of our knowledge, the first to extract context-dependent semantic substitution networks - that is, self-referential representations of language - from existing texts.

\section{Method}
\subsection{Notation and language model}

Our object of analysis is a corpus of text $D$, a collection of sequences of the form $s=(s_{1},s_{2},\ldots,s_{i}, \ldots, s_{m})$. Sequences consist of tokens - that is, words - from the vocabulary $\Omega$, such that $s_{i} = \mu \in \Omega$. For a given $s_{i}$, we denote the remaining elements of the sequence as $s_{-i}=(s_{1},s_{2},\ldots,s_{i-1},s_{i+1},\ldots)$.
For example, such a sequence may be 
$s=(The, founder, is, a, leader)$ where $s_2=founder$.\footnote{In the empirical example that follows, we define a sequence as a single sentence. Increasing the sequence length improves the resolution of the semantic analysis at the cost of additional computational complexity.}

We train a language model $P_D$ on $D$ using the Masked-Language-Modeling task detailed in \cite{devlinBERTPretrainingDeep2018}. Next, we extract its output as a probability distribution over the vocabulary $\Omega$. Specifically, the model $P_D$ predicts the token $s_i$ in a sequence $s$, taking the other tokens $s_{-i}$ as input. The random variable corresponding to this prediction is denoted as $w_{i}$. That is, we derive $P_D(w_{i}|s_{-i})$ such that if $s_i=\tau$, then $P_D(w_{i}=\tau|s_{-i})$ is maximized. Whenever we write $s_i$, we mean the realization found in the data, whereas $w_i$ is the random element as estimated by the probability model $P_D$.

In the example above, the ground truth is $s_2=\tau=founder$, and the model predicts $w_{i}$ given the sequence $s_{-i}=(The, - , is, a, leader)$. 

In general, a language model can predict a missing word in any input sentence $s$. Since our focus is on the analysis of existing text, we know the identity of every $s_i$ for $s\in D$. For example, we would be aware that $s_2=founder$. As such, we write $P_D(w_{i}|s_{-i}, s_i=\tau)$ noting that $s_i$ is not an input of the model, but rather a fact derived from the corpus $D$. Crucially, the random variable $w_i$ is distinct from its realization $s_i=\tau$ in the corpus. $P_D$ is an estimate of a latent data-generating process on the basis of the sample $D$ (see appendix \ref{app:ch1:scope}).

The language model’s input is a single sequence, whereas the model’s training uses the entirety or a subset of the corpus $D$. We make two conceptual divisions of $D$. 

First, we define a 
\textit{context} as subset $C \subseteq D$. For example, $C=\{s_1,s_2,\ldots\}$ may include all sentences that are about start-ups or all sentences that contain the word "founder". Inferences we make in $C$ reveal specific linguistic properties of $C$ in relation to all sequences of $D$ from which the model is trained.

Second, we compare the language between distinct corpora, say $D_1$ and $D_2 \subseteq D$. We then estimate a separate, statistically independent model $P_{D_1}$ and $P_{D_2}$ for each corpus. Whether we analyze two sets of sequences as distinct corpora (i.e., $D_1$ and $D_2$) or as two contexts within a larger corpus ($C_1$ and $C_2$ contained in $D$) depends on our assumptions about the generating processes. Generally, a split into separate corpora is preferable, insofar as $D_1$ and $D_2$ can be considered separate texts between which there may have been a fundamental shift in language. Entire issues of a publication or complete books would, for example, would constitute independent corpora. By contrast, individual chapters of a book would be considered context.
Note, however, that our method allows us to aggregate both across contexts and corpora, as the output of the language model remains in the same space: the simplex across a sufficiently broad lexicon of words.

\subsection{Lexical substitutes and word senses}

Prior literature has shown (\cite{hewittStructuralProbeFinding2019,tenney_bert_2019}) that transformer-based language models such as \textit{BERT} accomplish their predictions by learning semantic and syntactic relationships.
BERT is trained to predict $P_D(w_{i}=\tau|s_{-i}, s_i=\tau)$ (\cite{devlinBERTPretrainingDeep2018}). To do so, it learns a \textit{set} of high-dimensional representations of tokens \textit{and} their context. These representations imply a large number of spatial positions and proximities between tokens which, in turn, allow the prediction of the probability vector $P_D(w_{i}|s_{-i}, s_i=\tau)$. Due to the continuity of these representations the probability vector is well-defined not only for the ground truth element $\tau$, but also for alternative tokens $\mu \neq \tau$.\footnote{We can further saturate these predictive distributions by introducing a minor modification to the model, as detailed in appendix \ref{app:ch1:saturation_expansion}.}  $P_D(w_{i}=\mu|s_{-i}, s_i=\tau)$ attains higher values insofar as $\mu$ constitutes an appropriate prediction for $s_i$ in terms of semantics, syntax and position within the sequential context. 

\begin{figure}[h]
    \centering
    \efbox{\includegraphics[scale=0.8]{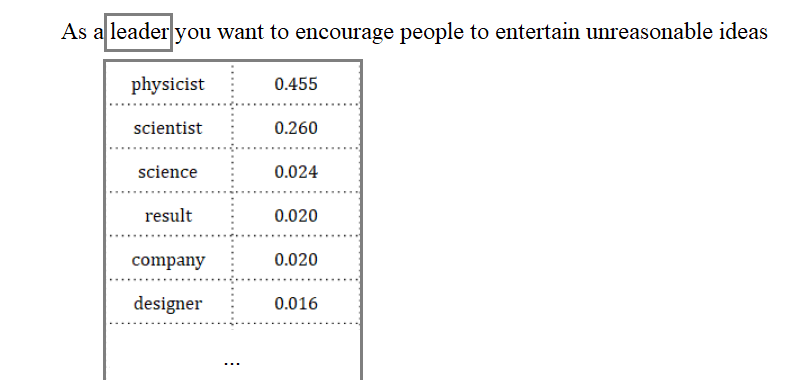}}
    \caption[Example of substitutions in a sentence]{Possible substitutes and associated probabilities $P(w_i|s_i=\tau, s_{-i})$ of $\tau=$"leader"}
    \label{fig:c1:reg_vectors}
\end{figure}

Given a sequence $s$ and a focal position $s_i$ with associated token $s_i=\tau$, we extract $P_D(w_{i}|s_{-i}, s_i=\tau)$. In the appendix, we detail how we transform this measure to be positionally independent, while retaining its probabilistic interpretation. Then the probability vector, $P_D(w_{\tau}|\tau \in s)$ gives alternative words that are appropriate to substitute for $\tau$ as it occurs in $s$. 
Figure \ref{fig:c1:reg_vectors} shows the substitute probabilities of the word $\tau=$"leader" in a sentence from our corpus.
\efboxsetup{linecolor=black,linewidth=1pt}

These \textit{lexical substitutions} reveal semantic information about $\tau$: to the degree that $\mu$ can replace $\tau$ (such that $P(w_i=\mu|s_i=\tau, s_{-i})\rightarrow1$), we can say that $\tau$ is used \textit{in the sense of $\mu$}. If, on the other hand, further terms constitute appropriate substitutes for $\tau$, such that both $P(w_i=\mu|s_i=\tau, s_{-i})>0$ and $P(w_i=\rho|s_i=\tau, s_{-i})> 0$, then $\tau$ has been used in a sense compatible with either terms $\mu$ and $\rho$. 
In figure \ref{fig:c1:reg_vectors}, a leader who \textit{entertains unreasonable ideas} invokes analogies along the idea of scientific pursuit, the attainment of results, the concept of commercial organization, and so on. In other words, "leader" is used in a specific sense of which $P(w, \tau \in s|s)$ is a quantitative representation. 

For that reason, the substitute distribution $P(w, \tau \in s|s)$ is said to define the \textit{word sense} of $\tau$ and an understanding of how $\tau$ is employed in the corpus at hand can be gained by analyzing all such substitute distributions in the corpus $D$ (\cite{alagicRepresentingWordMeaning2021,amramiBetterSubstitutionbasedWord2019}). 
Note, however, that this procedure is only feasible for words that occur frequently.\footnote{This is not a limitation of the language model. Indeed, BERT draws upon higher-order connections in the text to learn semantic information about words that do not occur frequently. For example, even if there are few sentences relating leaders to scientific pursuits, BERT could still learn that leaders are \textit{in a sense} similar to scientists, since both are related to \textit{results}, \textit{people} and \textit{ideas} in a similar way.} Indeed, the \textit{sense} in which "leader" is used is a continuous and highly varying concept.
In contrast to earlier models, most of the variations in these substitute predictions from BERT arise from context ($95\%$, \cite{ethayarajhHowContextualAre2019}). That is, substitute relationships, and thus word senses, differ for each unique sentence in which a word like "leader" is invoked: $P(w, \tau \in s|s=s_1)$ will differ from $P(w, \tau \in s|s=s_2)$, which will differ from $P(w, \tau \in s|s=s_3)$ and so on. Typically, there are as many word senses as there are occurrences. As a consequence, the analysis requires several additional techniques, such as clustering methods (\cite{amramiBetterSubstitutionbasedWord2019}); given the large contextual variation, it also requires a significant number of occurrences of the focal term.

Crucially,  our understanding of the word sense of the word "leader" depends on the semantics of terms that define the \textit{word sense} of the occurrence. For example, what is the sense of \textit{science} given by a particular combination of scientific terms that substitute for "leader"? If "leader" appears in many different senses of meaning, is the same not true for words like \textit{scientist}?
Substitute semantics represent a self-referential sense of meaning. "Leader" is written in the sense of a scientist, but the sense of "scientist" is equally dependent on the contextual and temporal semantic properties of the corpus. 

Generally, this circularity is a “overriding problem in semantic analysis” (\cite{goddard2010semantic}) which can be solved in one of two ways. Either, one brings in an (exogenous) set of primitives, or one specifies semantics purely as a system of relationships (\cite{goddard2010semantic}).
Taking the second option, we seek to analyze the semantics of a  text without primitives such as  predetermined semantic labels or categorizations. Hence, we need to account for the interdependence of meaning, taking the word sense of one occurrence as evidence for the word sense of another. That is, we need to conceptualize lexical substitutes as relationships between words conditional on context. In the next section, we develop such a representation.

\subsection{Lexical substitutes as relationships}

For a given sequence $s$ and an occurring word $\tau$, the distribution $P_D(w_{\tau}|\tau \in s)$ defines \textit{a set of relations between $\tau$ and other tokens.}, which we denote as a function of the input sequence $s$ and the ground truth element $\tau$. Using the distribution $P_D(w_{\tau}|\tau \in s)$, we define the dyadic measure of the probability that 
$\tau$ occurs in $s$, and $\mu$ is an adequate substitute. That is
$$
g_{\mu,\tau}(s) \defeq P(w_{\tau}=\mu|s)
$$

\begin{figure}[h]
    \centering
\efbox{
\begin{tikzpicture}[node distance={40mm}, thick, main/.style = {draw, circle}]
\node[main] (1) {\small{$\tau=$leader}}; 
\node[main] (2) [left of=1] {\small{$\mu=$scientist}}; 
\node[main] (3) [above of=2] {\small{$\rho=$physicist}}; 
\node[main] (5) [right of=1] {\small{...}}; 
\node[main] (4) [above of=5] {\small{$\gamma=$science}}; 

\draw [->] (3) to node [sloped, above] {\small{$g_{\rho,\tau}(s)=0.455$}} (1);
\draw [->] (2) to node [above] {\small{$g_{\mu,\tau}(s)=0.260$}} (1);
\draw [->] (4) to node [sloped, above] {\small{$g_{\gamma,\tau}(s)=0.0.024$}} (1);
\draw [->] (5) to node [above] {\small{$g_{\ldots,\tau}(s)$}} (1);
\end{tikzpicture}}
\captionsetup{justification=centering}
    \caption[Substitution ties from example sentence]{\small Substitute ties arising from example sentence in figure \ref{fig:c1:reg_vectors}}
    \label{fig:c1:sub_ties_example}
\end{figure}
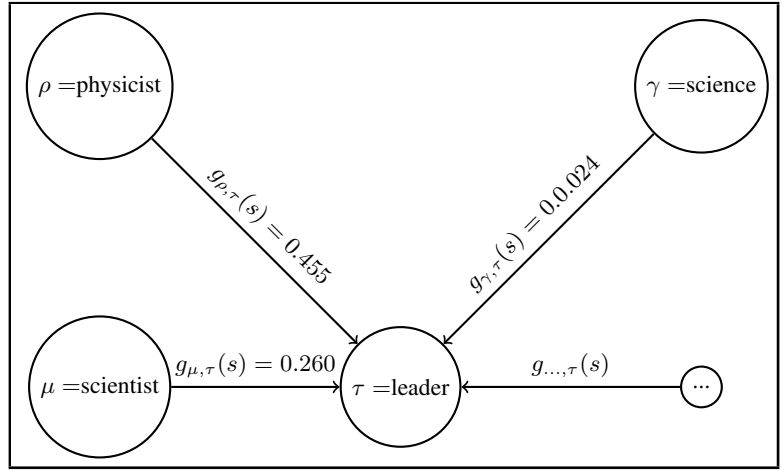

Figure \ref{fig:c1:sub_ties_example} illustrates the substitute relationships arising from the occurrence of $\tau=$"leader" as shown in figure \ref{fig:c1:reg_vectors}.

Note that the individual values of $g_{\mu,\tau}(s)$ can be understood from both directions: First, $\tau$ is being substituted by $\mu$ and, second, $\mu$ \textit{is substituting} for $\tau$. In the former direction, $\tau$ is the focal token. This direction represents the word sense of a particular occurrence of $\tau$ as mentioned above. In the latter direction, $\mu$ is the token of interest. Nevertheless, conceiving these probability values as dyadic relations ensures that both directions are taken into account.

In turn, a high value of $g_{\mu,\tau}(s)$ implies that $\mu$ is one element constituting the sense in which $\tau$ is invoked. In other words, $\mu$ transmits meaning toward $\tau$. 
Consequently, the collection of relationships $g_{\mu,-}(s_1), g_{\mu,-}(s_2), \ldots$ represent the meaning that $\mu$ can transmit to any alter terms and constitute the semantic composition (or simply the semantics) of $\mu$ in this context. For example, in the context of a company, the term $\mu=$"founder" may substitute for words such as "CEO" or "executive". In the context of a political movement, the semantics of "founder" and, thus, the terms it substitutes for, would differ.

The advantage of the proposed relational model is threefold. 
First, we can flexibly examine the semantics of $\mu$ - the meaning it may transmit - \textit{for any set of sequences}, whether $\mu$ occurs in them or not. We also find improved recall for $\mu$ replacing alters, but no loss of precision: substitute relationships exist even for tokens $\tau$ that occur very infrequently. 

Second, the relational model allows for a recursive semantic definition of each alter term and therefore structural semantic analyses, as we will detail.

Finally, \textit{we can identify the contexts} in which $\mu$ has semantic relationships to alter tokens and consider relationships that fit to a desired set of sequences. To this end, we first expand our formal model of substitute ties to allow for arbitrary sets of contexts.

\subsection{Aggregation of ties into a network}

We now develop operations to aggregate sequence or occurrence based information across diverse contexts. 
The structure arising from substitution ties of a given occurrence is a graph where tokens like $\mu$ and $\tau$ are nodes or vertices, and $g_{\mu,\tau}(s)$ are directed and weighted ties between them. Recall that Figure \ref{fig:c1:sub_ties_example} shows a snapshot of such a graph arising from a single occurrence in a single sentence $s$. 

Similarly, for a collection of such sequences, the only informational structure that incurs no loss in relational information remains a graph. Indeed, since there are many sequences in the corpus, the relations span a multi-graph - a network with parallel edges - indexed by $s$. There are infinitely many possible sequences, such that the language model is informationally identical to a multigraph with infinitely many parallel edges. Our sample, that is, the model's restriction on the corpus $D$, is informationally equivalent to the multi-graph we derive by querying all sequences in the corpus $D$.

For any context, that is, for any collection of sequences $C$, we can aggregate these substitute ties and, in turn, reduce the multi-graph to a simple directed and weighted graph representing language in a single context $C$.

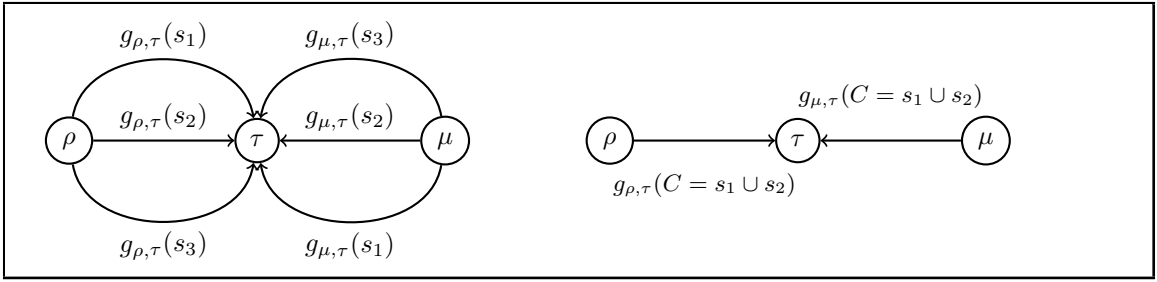
\begin{figure}[h]
    \centering
\efbox{
\begin{minipage}{0.45\textwidth}
\begin{tikzpicture}[node distance={25mm}, thick, main/.style = {draw, circle},xscale=0.9,yscale=1] 
\node[main] (1) {$\tau$}; 
\node[main] (2) [right of=1] {$\mu$}; 
\node[main] (3) [left of=1] {$\rho$}; 
\draw [bend right=800,<-] (1) to node [sloped, below] {$g_{\mu,\tau}(s_1)$} (2);
\draw [bend right=-800,<-]  (1) to node [sloped, above] {$g_{\mu,\tau}(s_3)$} (2);
\draw [<-] (1) to node [sloped, above] {$g_{\mu,\tau}(s_2)$} (2);
\draw [bend right=800,<-]  (1) to node [sloped, above] {$g_{\rho,\tau}(s_1)$} (3);
\draw [bend right=-800,<-]  (1) to node [sloped, below] {$g_{\rho,\tau}(s_3)$} (3);
\draw [<-] (1) to node [sloped, above] {$g_{\rho,\tau}(s_2)$} (3);
\path[step=1.0,black,thin,xshift=0.5cm,yshift=0.5cm] (2,0) grid (-4,1);
\end{tikzpicture}
\end{minipage}
\hfill
\begin{minipage}{0.45\textwidth}
\begin{tikzpicture}[node distance={25mm}, thick, main/.style = {draw, circle},xscale=0.9,yscale=1] 
\path[step=1.0,black,thin,xshift=0.5cm,yshift=0.5cm] (-2,-2) grid (1,1);
\node[main] (1) {$\tau$}; 
\node[main] (2) [right of=1] {$\mu$}; 
\node[main] (3) [left of=1] {$\rho$}; 
\draw [<-] (1) to node [sloped, yshift=0.3cm,above,font=\small] {$g_{\mu,\tau}(C=s_1 \cup s_2)$} (2);
\draw [<-] (1) to node [sloped, yshift=-0.3cm,below,font=\small] {$g_{\rho,\tau}(C=s_1 \cup s_2)$} (3);
\end{tikzpicture}
\end{minipage}}
\captionsetup{justification=centering}
    \caption[Aggregation of substitution ties]{\small Aggregation of substitution ties across contexts.\\ Left: Incoming ties of $\tau$ in multigraph $G$ \\ Right: $G^C$ with $C=s_1 \cup s_2$}
    \label{fig:c1:agg_ties}
\end{figure}

In appendix \ref{app:ch1:aggregation}, we show that we can estimate the latent probabilities of a given sequence from the corpus, and use basic rules of probability to aggregate $g_{\mu,\tau}(s)$ across a context $C$, leading to

$$g_{\mu,\tau}(C) = P_D(w_{\tau}=\mu| s \in C)$$

Fixing $C$, for example, as the as the set of sentences that include the word "company", defines a simple, directed graph with ties $g_{\mu,\tau}(C)$ that measure how likely $\mu$ is to replace $\tau$ in $C$. In other words, to the degree of $g_{\mu,\tau}(C)$, $\tau$ is used in the sense of $\mu$ in the context $C$. 
We denote this graph by $G^C$. Since $G^C$ is based on lexical substitutions, it is a network of semantic relations. Nevertheless, we call $G^C$ a \textit{semantic substitution network} in order to differentiate the similarity-based notion of semantics from the conventional meaning of semantic networks (e.g., \cite{miller1995wordnet,qi2019openhownet}).
Figure \ref{fig:c1:agg_ties} gives a schematic overview of an aggregation from $G$ to $G^C$.

\section{Aspects of meaning in the substitution network}
Lexical substitution semantics are contextual and self-referential. 
For example, if we were asked to describe the meaning of the word "founder", we would do so by referring to other words in terms of their semantic aspects on the one hand, and their associated contexts on the other: A "founder" is a \textit{creator of a company}, but - as an another semantic aspect - a "founder" could also be \textit{the head of the firm's formal hierarchy}. A founder is typically not the \textit{creator of a piece of art}. In this way, the meaning of a word or a sequence of words can be defined as a set of abstract analogy relations (\textit{founder-creator}), each of which depends on a context (\textit{... of a company}). 
Substitution relations are the \textit{operationalized notions of these analogies as inferred by the language model.}
That is, if our language model judges a token to be an adequate substitutes, this is because authors have used these words in a similar way relative to other words in the text.

In appendix \ref{app:ch1:semantic_structures}, we further detail the meaning conferred by substitution ties.
In the next section, we describe how the substitution network can identify three aspects of meaning: substantive, contextual, and structural, which we will discuss in turn.\footnote{In appendix \ref{app:ch1:structural_identification} we show how we can use the different stages of aggregation to disambiguate syntactic from semantic relationships.} 

\subsection{Substantive aspects of meaning}

The semantic substance of a word is the meaning it confers to other words. For a given context $C$, the substantive aspect of meaning of a word $\mu$ can be identified from the substitution relations between $\mu$, its neighbors, and higher-order relationships in $G^C$.

Consider two words in $G^C$ and assume that the language model has learned that both words are appropriate substitutes. These substitution ties are equivalent to \textit{synonymy}, where, however, such synonymy ranges from \textit{near} to \textit{strong}. Strong synonyms are identical in meaning while near synonyms are words that merely have a \textit{similar} semantic and syntactic function.

Near synonymy implies and is implied by substitute relationships: If one word appears with a meaning shared by the other word, then a well-trained language model would return a high likelihood of substitution.
However, this substitution tie, even if it has a high value, does not imply \textit{strong synonymy}: even $g_{\mu,\tau}(\ldots)=1$ in $C$ does not allow the conclusion that both terms are semantically identical. This is so, because the meaning of the occurring word may include nuances that the other term lacks, or its meaning in the given sentence may be ambiguous. Finally, in other sentences in $C$, the two words may not even be synonyms at all.

We claim that the strength of synonymy has structural consequences for the semantic network. If the meaning of two words were truly indistinguishable, then their use would not differ systematically in the corpus.\footnote{More precisely: if we observe systematic differences in use, we can not infer that the two words have an identical meaning in the language internal to the given corpus.} Consequently, both words would not only be pairwise substitutes, they would also share ties of similar weight to the same set of other tokens in $G^C$. In particular, if one token is a substitution for another third, then so should be the other. By contrast, if either word connotes different, our procedure would show variation in the ties associated with alter tokens.

Taken together,  the \textit{substantive semantic identity} of a word is given by the collection of substitution ties, or near-synonymy relations, towards other words. Words have the same substantive semantic identity, and are therefore indistinguishable strong synonyms \textit{in $C$}, iff they share the same near-synonymy relationships to alter terms. As no two words are truly identical, we generally measure the degree to which \textit{substantive semantic identities} overlap - the degree to which two words are structurally equivalent in $G^C$.

For example, we find that the terms "founder" and "CEO" are near synonyms in the language used in \textit{Harvard Business Review}. They are pairwise substitutes in many sequences in which they appear.
\begin{center}
\begin{table}[ht]
    \centering\scriptsize
 \begin{tabular}{||c| c| c||}
 \hline
 \textbf{"founder" as in ...} & \textbf{"founder" and "CEO" as in ...} &  \textbf{"CEO" as in ...} \\ [0.5ex]
 \hline\hline
CEO	& president	&board \\ \hline
founders	& chair	&executive \\ \hline
chairman	& director	&manager \\ \hline
cofounder	& entrepreneur	&leader \\ \hline
author	& head	&company \\ \hline
insider	& owner	&founder \\ \hline
engineering	& customer	&boss \\ \hline
owner	& consultant	&CEOs \\ \hline
MBA		&&management \\ \hline
vice		&&top \\ \hline
founding		&&person \\ \hline
degree		&&team \\ \hline
vc		&&employee \\ \hline
\hline
\end{tabular}
    \caption[Alter tokens proximate to "founder" and / or "CEO"]{\small Alter tokens proximate to "founder" and "CEO", measured by their relative contribution to the total mass of substitute ties $\frac{g_{\mu,\tau}(C)}{\sum_{\rho} g_{\mu,\rho}(C) }$. Left column: Tokens more proximate to "founder" than "CEO". Right column: Tokens more proximate to "CEO" than  "founder". Both columns ranked by differences between substitution tie strength. Middle column: Tokens proximate to both "founder" and "CEO", ordered by similarity of tie strengths. $G$ aggregated across the entire corpus of \textit{Harvard Business Review} from 1980 to 2020.}
    \label{tbl:c1:"CEO"_founder_syn}
\end{table}
\end{center}

However, table \ref{tbl:c1:"CEO"_founder_syn} illustrates that both words have a distinct semantic identity and are only \textit{structurally equivalent} to each other relative to a small subset of alternative words, which are shown in the middle column of the table. On the other hand, the left column shows alter terms which are more strongly associated with "founder" than with "CEO", for example "cofounder", "owner" or "engineering". Similarly, terms like "board" or "executive" are more likely to be substitutable by "CEO" than "founder". 

\subsection{Structural aspects of meaning}

The prior section asserts that the strength of synonymy between two words equals the similarity of their substitution relationships to all words. The reverse of this assertion is that strong synonyms are  indistinguishable in their use in the corpus. In other words, the more substitution relationships of two words overlap, the less variation there is in semantics, and the less distinct information about meaning we can learn if one word appears in place of the other.

We call a set of words that convey similar meaning a \textit{semantic cluster}. These words share stronger substitution ties among themselves, compared to words outside the cluster. We can employ a number of standard clustering algorithms (e.g., \cite{traagLouvainLeidenGuaranteeing2019a,blondelFastUnfoldingCommunities2008,rosvallMapEquation2009,rosvallMapsRandomWalks2008}) to derive the groupings representing semantic clusters in $G^C$.

Since the degree of strong synonymy is a continuous measure, these clusters arise in $G^C$ hierarchically. This enables us to structure the ego network of a focal token $\mu$ and the semantic clusters to which it relates (\cite{blondelFastUnfoldingCommunities2008,rosvallMapEquation2009}). We find that top-level clusters identify the \textit{functional aspects} - the parts of speech - in which $\mu$ appears in the text (see appendix \ref{app:ch1:structural_identification}). Within such a top-level "syntactic" cluster, we then identify communities of words that share substantive meaning, and denote them as $S_k$.  For example, in the ego network of the term "founder", we find a cluster of roles, which itself is comprised of semantic subgroups like  $S_i=\{$\textit{"owner", "entrepreneur", "pioneer", "innovator", "designer", \ldots}$\}$, or $S_j=\{$\textit{"professor", "dean", "researcher", "student", \ldots}$\}$.

The average strength of substitution ties to and from $\mu$ measures the semantic relation of the focal word to these clusters. While $\mu$'s meaning is characterized by relations to individual words, its average relation to semantic complexes gives a higher level view of the aspects of meaning that arise from its use in the text. We define

$$ g_{\mu, S_j}(C)=\frac{1}{|S_j|}\sum_{\tau \in S_j} g_{\mu, \tau}(C),$$

and we define $g_{ S_j,\mu}(C)$ symmetrically.\footnote{Since both the size of semantic complexes and the distribution of substitution ties within such a complex may be highly skewed, we alternatively define a weighted average as
$$ g^*_{\mu, S_j}(C)=\sum_{\tau \in S_j} w_{S_j}(\tau,C) g_{\mu, \tau}(C)$$
where $w_{S_j}(\tau,C)=\frac{g_{\mu, \tau}(C)}{\sum_{\tau \in S_j} g_{\mu, \tau}(C)}$.}
More importantly, the granularity of this composition is variable. We can further divide these clusters by \textit{increasing the stringency of our definition of strong synonymy}. The finest-grained groupings in the semantic substitution network of roles associated with "founder", for example, are singular terms or dyads like \textit{"owner"-"proprietor"}, \textit{"consultant"-"analyst"}, or \textit{"pioneer"-"visionary"}. \textit{"husband"-"wife"}, \textit{"candidate"-"applicant"}, or \textit{"colleague"-"friend"}.

\efboxsetup{linecolor=black,linewidth=1pt}
\begin{figure}[h]
    \centering
    \efbox{\includegraphics[width=0.9\textwidth]{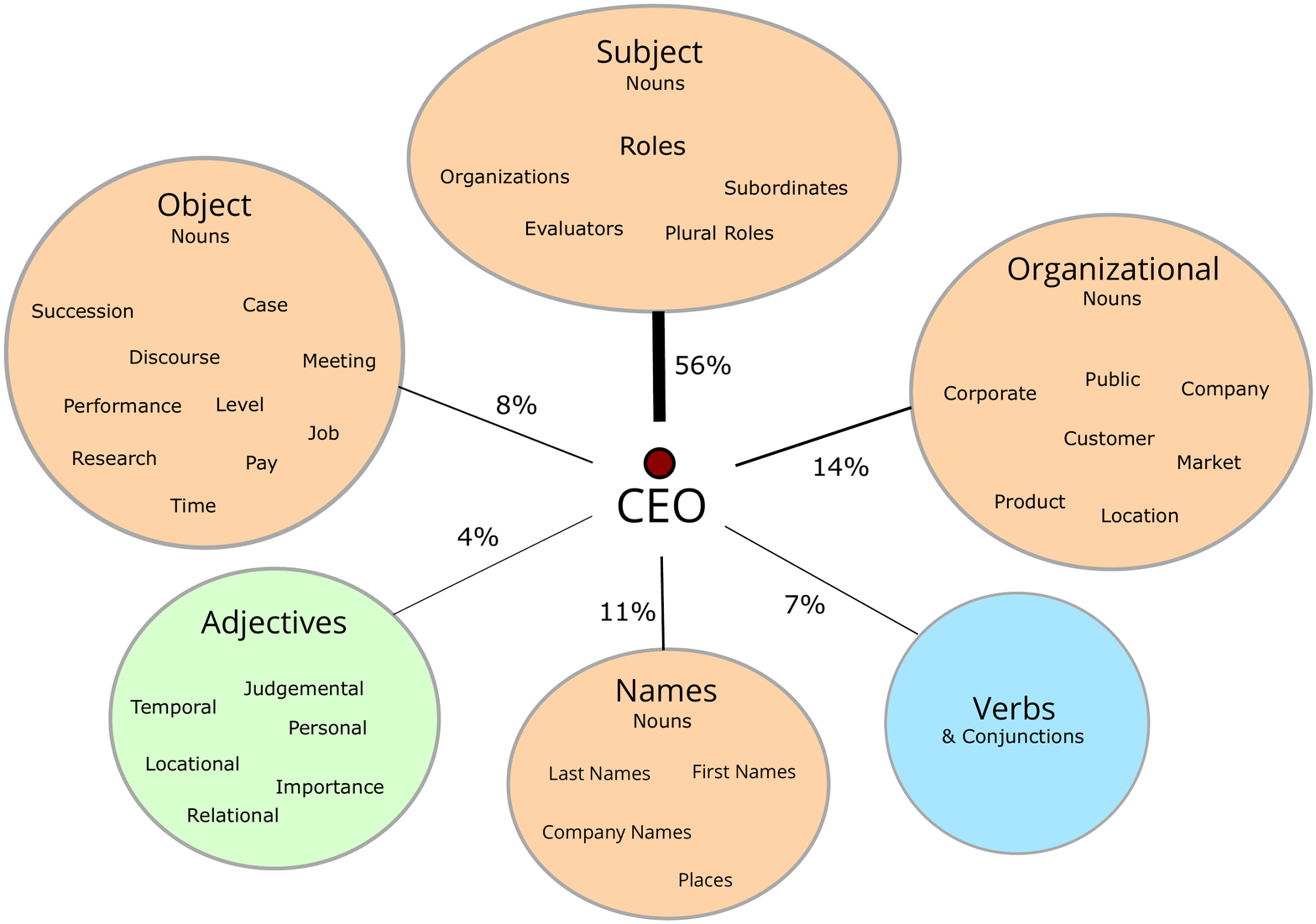}}
    \caption[Syntactic and semantic composition for "CEO"]{Syntactic and semantic composition for "CEO". Large circles show the first-level clusters and their part-of-speech. Percentages denote $g_{\tau, S_j}(D)$ relative to $\sum_{S_j} g_{\tau, S_j}(D)$. In each circle, sub-clusters comprised of more than $40$ words are shown, except for the Verb cluster. Names correspond to the most proximate token or the semantic category of the sub-cluster.}
    \label{fig:c1:syntacticcluster}
\end{figure}
Figure \ref{fig:c1:syntacticcluster} depicts a schematic illustration of the ego-network of "CEO" and its first-level clusters, their relative proximity to the focal term, as well as a description of the second level clusters.

\subsubsection{Semantic importance and breadth}

Having focused on the immediate substitution relations of a word, we now turn towards its position in the semantic structure of language. Here, we can broadly associate two additional semantic properties with a word: \textit{importance} and \textit{breadth}.\footnote{See the detailed discussion in appendix \ref{app:ch1:centrality}.}

A word has a high semantic importance, if it contributes meaning to other words, in particular those words that are themselves integral to the understanding of a given sentence. For example, at some point in time, the word "man" may constitute an important semantic aspect of many important organizational roles like leader, CEO or president. The use of these roles then defines the meaning a reader derives from, for example, consuming financial news in the popular press. The word "man" is central, in the sense that it constitutes a crucial, perhaps taken-for-granted understanding of what is said in these texts referring to such roles. In contrast, other times, the association of organizational roles and terms like "man" relative to terms like \textit{woman} may be more balanced. In that case, \textit{man} would be less central, and the meaning of texts invoking such organizational roles would be colored less by the element of masculinity. We can measure semantic importance using network centrality (\cite{bonacich1987power,newmanMeasureBetweennessCentrality2005,page1999pagerank}).

A different element of the structural aspect of semantics if its \textit{breadth.} In a given context, say, texts about a company's marketing strategy, a focal word such as "advertisement" has a relatively narrow meaning. It may appear in many sentences, and it may even be a likely substitution for many important words in the text. However, whenever it is a substitute, the meaning that it gives the occurring term is relatively clear. Structurally, terms that "advertisement" can replace, are also likely substitutes for each other. By contrast, a term like "relations" has a broader set of semantics. It may substitute for words in sentences about public relations, customer relations, competitor relations or internal relations. We might find that across these sub-contexts, replaced terms share only weak semantic ties. Thus, "relations" has greater semantic breadth: it contributes meaning, here an aspect of relatedness, to several disconnected conversations. Since "relations" would have a brokerage position between semantic clusters in our network, we measure semantic breadth with  betweenness centrality (\cite{everettEgoNetworkBetweenness2005,newmanMeasureBetweennessCentrality2005}).

Both importance and breadth are typically correlated, but this correlation need not hold for every word. A word with high importance and narrow breadth gives a precise sense of meaning. On the other hand, a word may be used broadly, but have little semantic importance. Such a word would have a vague meaning.


In contrast to earlier approaches, deep learning models are context sensitive and return different semantic relationships depending on the sequence under consideration. In the next section, we detail how this ability can sharpen our understanding of the semantic identity of a focal token.

\subsection{Contextual aspects of meaning}

The method introduced up to this point has specified a context $C$ as any subset of sequences. Measures and inferences are conditional on context, because of the high  semantic variability between different sequences.
In practice, the contextual divisions need to be determined by the researcher. One particular choice of context is to set $C=D$. In this case, the present approach is similar to co-occurrence analyses (\cite{weedsCooccurrenceRetrievalFlexible2005}) or shallow word embeddings (\cite{mikolovEfficientEstimationWord2013}), albeit with the improved resolution and precision allowed by deep language models. 
For example, consider a set of corpora for each year of publication of a newspaper, say $D_{1990}$ to $D_{2000}$. We can then identify the most prominent semantic complexes in terms of $g^*_{\tau, S_j}(C)$ for each year and compare them in sequence, setting $C=D_{1990}$, $C=D_{1991}$ and so on. This allows an assessment of how the meaning of $\tau$ has changed across these years. We can also compare these yearly measures to the semantic substitution network derived across all years, that is, $C= D_{1990} \cup D_{1991} \cup\ldots$ to identify significant changes in the meaning, or use this overall network to derive intra-corpus deviations for further analyses.

However, the largest advantage of contextual models is that they can be used to \textit{identify relevant contexts}. Recall that a "CEO" is \textit{a leader of a company}, but not a \textit{leader of a country}. As this example shows, semantic ties depend crucially on the context they are embedded in. 

We present two measures. First, we can use our network to identify the contexts in which dyads of substitutes are likely to occur. That is, when a semantic complex identifies an aspect of meaning, we can inquire in which contexts this aspect of meaning is relevant.
Doing so is straightforward in the multigraph $G$ that is indexed by sequences. Consider a dyad $\tau$ and $\mu$. We derive a measure of its context by examining the substitute distribution of other words in each sequence $s$ in which $\tau$ and $\mu$ are substitutes. In appendix \ref{app:ch1:dyadic_context}, we introduce the \textit{dyadic context} $\underline{q}^{\mu,\tau}_{\rho}(C)$ for a given dyad $\mu, \tau$ as a distribution of contextual words $\rho \in \Omega$. If $\underline{q}^{\mu,\tau}_{\rho}(C)$ is high for some word $\rho$, then $\rho$ is a likely substitute for \textit{other words} in sentences in which $\tau$ is used in the sense of $\mu$. That is, $\rho$ contributes to the substantive meaning of the context in which $\tau$ and $\mu$ are semantically related.

Our second measure defines the context in which a focal token is a substute for, or transmits semantic content to, textit{any} other term. That is, we seek to specify the elements of meaning that a word provides, by examining the contexts in which such transmissions take place. Operationally, the measure is similar to $\underline{q}^{\mu,\tau}_{\rho}(C)$. However, instead of focusing on a specific dyad of substitution, we aggregate across such dyads. Let 
$$
\underline{q}_{\mu}(\rho)=\underline{q}_{\mu, \rho}=\sum_{\tau} \underline{q}^{\mu,\tau}_{\rho}(C)
$$
which defines a network $Q$ of pairwise relations $\underline{q}_{\mu,\rho}$. Two words $\mu$ and $\rho$ are neighbors in $Q$, if $\rho$ is a defining element of the context in which $\mu$ replaces other words. That is, if words are used in the sense of $\mu$, then $\rho$ specifies the context of this meaning. For example, if the usage of "leader" is \textit{leader, as in the manager of a company}, then $\mu$ corresponds to the meaning-giving term "manager", while $\rho$ specifies substitution to occur in the context of a \textit{company}.

Thus, for a focal token, $\underline{q}_{\mu}(\rho)$ defines a set of other words that describe the context in which it is a relevant substitution. $\underline{q}_{\mu}(\rho)$ therefore serves to define and cluster the different context sets $C$. Given a focal token $\mu$, we consider only those sequences in which an alter token $\rho$ is a likely substitution for \textit{any position}. Alternatively, similar to earlier co-occurrence methods, we can define the contexts as those sequences in which an alter token appears. The latter definition considers only considers sentences in which these contextual terms occur, whereas the former definition uses the language model's substitutions of a sequence to judge whether the sentence should be part of the desired context set. Intuitively, assessments based on substitutes have a higher recall: more sequences aligned with the desired context can be found. Given the sophisticated representations generated by modern language models, this, however, need not come at a significant cost to precision. For that reason, we base our assessments on substitutes.

Furthermore, focusing on the dyad $\rho$ and $\mu$ allows us to interpret $\underline{q}_{\mu}(\rho)$ as a tie in a contextual network $Q$. In this network, outgoing ties from $\mu$ sum to unity and admit a probabilistic interpretation: Given that $\mu$ is a good fit in a focal sequence, what is the likelihood that $\rho$ appears in the sequence as well? Following the paths through $Q$ to a third token, say $\gamma$, consequently defines the likelihood that the triad of $\mu$,$\rho$ and $\gamma$ appear jointly. 
Note also that this relatively complex query is derived entirely from the language model trained on our corpus and yet, is directly available through our network $Q$. Indeed, paths in $Q$ therefore allow us to query conditional likelihoods of entire sequences, or parts of sequences, from the language model. While we do not pursue this point in the present work, it should be noted that this technique allows formally well-defined analyses of the language model's behavior that circumvent the issue of layer aggregation. Appendix \ref{app:ch1:context_element} discusses the implications in more detail.

\section{Illustrated Example: Founders in \textit{Harvard Business Review}}

\subsection{Introduction}

To illustrate our method, we examine an organizational role at core of any business: the founder. Founders, "those individuals involved in actualizing the steps of organizational founding” (\cite[p. 709]{nelson2003persistence}), are important in shaping the identity and evaluating the impact of businesses. Nevertheless, considerable difficulty arises in defining the role of a founder and the associated complex semantics (\cite[p. 708]{nelson2003persistence}).

The founder of a business inhabits the tangible position of creating, driving forward, and, in most cases, leading the organization. Success and failure of the firm are commonly associated with the persona of the founder (\cite{baron1999building}) and founders are seen as initiators and determinants for the future prosperity of the company (\cite{carland1984differentiating, gartner1985conceptual}). The role and identity of a founder could therefore be conceptualized by his or her entrepreneurial activities, and the founder's competency in the area of innovation and business development should be a strong predictor for organizational success.

By contrast, the empirical evidence as to the impact of the founder on the success of a business is mixed at best (\cite{adams2009understanding, barontini2006effect, bennedsen2007inside}). Several possible explanations exist. On the one hand, successful ventures are rarely created by "lone geniuses" but rather by multifaceted teams (\cite[p. 224]{klotz2014new}). As such, the actual role of any one founding member may differ from a definition focused on a single person. On the other hand, even a single founder inhabits a multitude of roles “salient [...] in her or his day-to-day work” (Powell and Baker, 2014, p. 1409). This association with different functions is highly contextual and dynamic. Indeed, "\textit{extant work has told us little about how founders' identities evolve (Crosina, 2018; Powell and Baker, 2017). Instead, we have seen research deliver a growing number of role and social identities [...] associated with founders.}” (\cite[p. 2]{o2020evolution}).

To demonstrate some aspects of our method, we analyze the roles and social identities associated "founder" in the entire corpus of \textit{Harvard Business Review}. Specifically, we ask
\begin{enumerate}
    \item What is the substantive semantic identity of a "founder" beyond its the creation of an enterprise?
    \item How has this semantic identity changed over time?
    \item What contexts are relevant for founders, and how does their semantic identity differ between them?
    \item Has the semantic importance of "founder" increased or decreased over time?
\end{enumerate}

In line with the prior technical exposition, we analyze the identity of the word "founder" by drawing direct analogies to other roles. These associations are based on higher-order semantics drawn from the descriptions of the behavior of individuals with the role of founder. The language model learns associations from identities and activities of founders in \textit{Harvard Business Review}, and estimates the similarity of these semantic associations to other roles - roles that convey similar meaning and serve a similar syntactic role.
Note that this approach differs from an analysis of literature \textit{about} founders. Such literature, based on the prevalent stereotypes of a founder, emphasizes the creative aspect of the role (\cite[p. 224]{klotz2014new}), perhaps in excess of what organizational reality warrants.

It bears mentioning that other approaches are available through context-aware semantic networks. For example, instead of focusing on analogous roles, one could examine the actions or behaviors that make a founder a fitting subject of a given sentence. Then, in turn, the semantics and context of such behaviors could be studied with our method. Sentiment and normative statements could further be used to characterize the different contexts in which substitution connections arise. While these details are of interest in their own right, they are also subsumed in the relationships between the word "founder" and other roles, on which we now focus.

\subsection{Data}

We extract all substitute ties from all articles of \textit{Harvard Business Review} from the years 1980 to 2020. This amounts to 1,112,589 sequences of length 3 or longer, and 10,780,603 usable word occurrences (excluding stop-words, errors, etc.). In this example, we retain 90$\%$ of the probability mass associated with each occurrence, leading to 319,000,000 relationships in our graph. This number  poses only a relatively mild computational demand on modern graph databases running on a single workstation computer.

\subsection{Substantive elements of the meaning of founder}

To illustrate the conceptual breadth of the role of a founder, consider a categorization of three prototypes of what a founder might be  (\cite{crosina2018becoming, powell2017beginning}). First, consider prototypical founders as innovators, inventors, visionaries, or entrepreneurs. 
Second, contrast these creative roles with those associated with organizational, managerial, and economic functions. This second prototype includes roles like president, director or manager that are situated within the organizational bureaucracy. Finally, the profile of a founder also includes roles like leader, colleague, boss or team member that represent a complex of social and coordinating activities necessary when developing a business.
To which degree does the meaning of "founder" in \textit{Harvard Business Review} conform to these prototypes?

We begin our analysis by examining the ego network of "founder" across all years from 1980 into 2020 and across all contexts. Recall that "founder" shares a substitution relation with another word, if that word has been used similarly, in similar sequences throughout \textit{Harvard Business Review}. Words that are syntactically similar and share a strong semantic relationship to in a given sentence, have thus been used in the sense of "founder" as defined in a given year.

There are numerous terms thus associated with "founder". However, any selection of words that are semantically related to "founder", also share semantic ties amongst themselves. To make use of this additional structure, we employ a hierarchical clustering algorithm to detect different hierarchies of communities within the ego network of "founder".
\efboxsetup{linecolor=black,linewidth=1pt}
\begin{figure}[h]
    \centering
    \efbox{\includegraphics[width=1\textwidth]{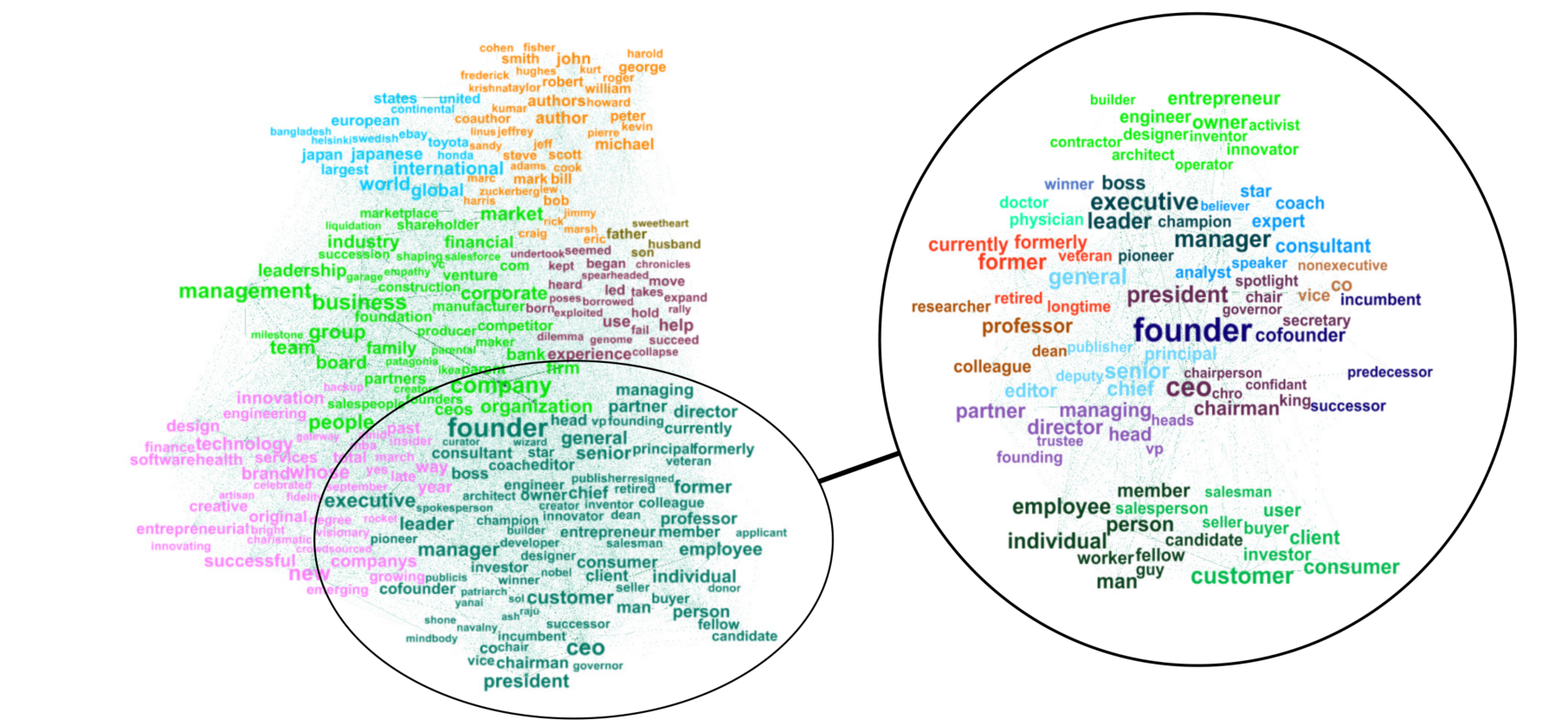}}
    \caption[Two levels of clusters for "founder"]{First-level clustering (left) and second-level clustering of roles associated with "founder" (right). Tokens with highest proximity to "founder" are shown in both cases.}
    \label{fig:c1:founder_clustering}
\end{figure}
On the left side of figure \ref{fig:c1:founder_clustering}, we show the words that "founder" was likely to replace in  \textit{HBR}. In this cross-sectional snapshot across all years from 1980 to 2020, colors indicate the communities of words identified by the clustering algorithm. As indicated by our theoretical discussion, this first level of clustering identifies syntactic-semantic relationships. In particular, "founder" is itself grouped in a cluster of roles. On the right  side of figure \ref{fig:c1:founder_clustering}, we focus on this cluster in the ego network of founder. Again, roles associated with founder are grouped into distinct communities, as indicated by their color. Since these roles are syntactically equivalent, the relationships are of semantic nature and thus represent one aspect of the semantic network of "founder": semantically related roles.

Table \ref{tbl:c1:roles} shows the most proximate clusters associated with "founder" on an even more granular level. We classified these semantic communities into the three aforementioned categories: administrative/hierarchical, creative/visionary, and social/coordinating.

\begin{figure}[h]
    \centering
    \efbox{\includegraphics[width=0.5\textwidth]{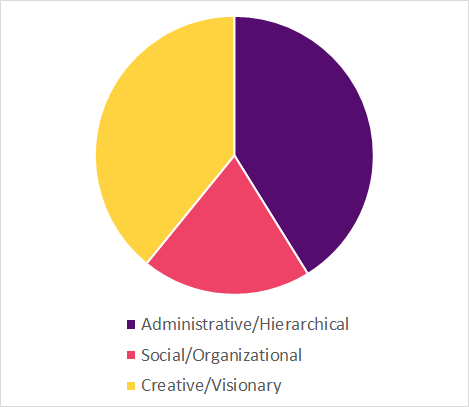}}
    \caption{Distribution of substitution ties of founder with categories of roles.}
    \label{fig:c1:role-dist}
\end{figure}

It is immediately apparent that the cluster around "CEO", "chairman" and "president" has a near synonymous relationship with founder. This relationship further shows very little variation across time or context. Roles in this cluster reflect the position of a founder across most of his or her tenure and display a similar semantic ambiguity. To reflect the fact that these roles also span our chosen categories, we do not take them into account when computing relative weights in table  \ref{tbl:c1:roles}.
\begin{center}
\begin{table}[ht]
    \centering\scriptsize
 \begin{tabular}{||c c c c c||}
 \hline
 \hline
 Cluster & Prominent Token & Average Tie Strength & Relative Weight & Classification \\ [0.5ex]
 \hline\hline
 CEO-chairman-president-chairperson-CHRO	&CEO	 &28.42 	&-	&- \\
\hline
cofounder-patriarch	&cofounder	 &7.59 	&17$\%$	&Administrative \\
\hline
founding-inception	&founding	 &4.11 	&9$\%$	&Creative \\
\hline
insider-outsider-newcomer	&insider	 &3.91 	&9$\%$	&Administrative \\
\hline
director-managing-partner-trustee-coordinator	&director	 &3.59 	&8$\%$	&Administrative \\
\hline
leader-executive-manager-ruler-officer	&leader	 &2.29 	&5$\%$	&Social \\
\hline
vice-nonexecutive-co-op-honorary	&vice	 &1.91 	&4$\%$	&Administrative \\
\hline
owner-proprietor-sponsor-organizer	&owner	 &1.85 	&4$\%$	&Administrative \\
\hline
head-VP-heads-face	&head	 &1.71 	&4$\%$	&Administrative \\
\hline
entrepreneur-innovator-investor-activist-thief	&entrepreneur	 &1.65 	&4$\%$	&Creative \\
\hline
editor	&editor	 &1.59 	&4$\%$	&Creative \\
\hline
chair-chairs	&chair	 &1.45 	&3$\%$	&Administrative \\
\hline
pioneer-guardian	&pioneer	 &1.12 	&3$\%$	&Creative \\
\hline
member-person-individual-adult-solo	&member	 &1.12 	&3$\%$	&Creative \\
\hline
father-mother-grandfather-uncle-dad	&father	 &1.04 	&2$\%$	&Social \\
\hline
boss-subordinate	&boss	 &0.80 	&2$\%$	&Social \\
\hline
employee-worker-trainee-expatriate	&employee	 &0.76 	&2$\%$	&Social \\
\hline
colleague-friend	&colleague	 &0.76 	&2$\%$	&Social \\
\hline
builder-architect-developer-contractor	&builder	 &0.76 	&2$\%$	&Creative \\
\hline
man-guy-woman-hillbilly-bitch	&man	 &0.73 	&2$\%$	&Social \\
\hline
professor-dean-researcher-student-instructor	&professor	 &0.72 	&2$\%$	&Creative \\
\hline
veteran-alumnus	&veteran	 &0.66 	&1$\%$	&Social \\
\hline
successor-heir	&successor	 &0.62 	&1$\%$	&Social \\
\hline
candidate-applicant-wits-finalist	&candidate	 &0.60 	&1$\%$	&Social \\
\hline
designer-engineer-operator-technician-stylist	&designer	 &0.52 	&1$\%$	&Creative \\
\hline
creator-avatar	&creator	 &0.48 	&1$\%$	&Creative \\
\hline
inventor-magnate-genius-titan	&inventor	 &0.46 	&1$\%$	&Creative \\
\hline
son-daughter-grandson-eldest	&son	 &0.39 	&1$\%$	&Social \\
\hline
visionary-narcissist-survivor-icon	&visionary	 &0.38 	&1$\%$	&Creative \\
\hline
consultant-broker-analyst-guru-banker	&consultant	 &0.37 	&1$\%$	&Creative \\
\hline

\hline
\end{tabular}
    \caption[Role clusters of "founder"]{Clusters semantically similar to "founder". Fifth level of clustering hierarchy from ego network of focal term. Relative weights are substitution weighted tie strengths $g^*_{\text{founder}, S_i}(D)$ relative to tie strength of all shown clusters except the "CEO" cluster. Only clusters top 30 clusters in terms of weighted proximity are included. }
    \label{tbl:c1:roles}
\end{table}
\end{center}
Instead, note that neither of the aforementioned prototypical ideas of a "founder" seems to be dominant in \textit{Harvard Business Review}.
Figure \ref{fig:c1:role-dist} illustrates that within the corpus of  \textit{HBR}, "founder" is just as likely to replace roles associated with administrative or hierarchical functions as roles classified as creative or visionary. Overall, the latter roles comprise less than forty percent of the "founder"'s semantic profile. Contrary to what is posited by much of the extant literature on founders, the organizational reality of a founder seems to be driven by administrative roles. 
\efboxsetup{linecolor=black,linewidth=1pt}

In the next sections, we analyze the change of the role profile of "founder" over the years in our sample. In the subsequent section, we extract the contexts in which a "founder" is a likely substitution for other words from the language model. Finally, we bring together associated roles with the contexts in which they semantically relate to "founder", and attempt to recover that multifaceted profile of roles that "founder" represents.

\subsection{Changes across years}
Changes in technology, institutional setting, globalization and also academic discourse are likely to transform the concept of "founder" over time. Indeed, eight of the ten most valuable companies in the world were founded during the interval from which our corpus is drawn. Furthermore, the founders of all but one of these companies have gained prominence during that time.\footnote{Apple Computer was created in 1977, and Saudi Aramco in 1933. Alphabet, Amazon, Microsoft, Tesla, TSMC, Facebook, Tencent and Alibaba were founded after 1980.} If the public's ideal of a founder was influenced by these personas, it is likely that the meaning of founder has changed in society. Similarly, when authors in  \textit{HBR}  analyze influential business leaders, these semantic connections also take on a new meaning as these founders and their companies grow and mature. Furthermore, roles associated with founders are themselves subject to change. For example, our prior analysis indicates that the most important social and organizational role for founders is to be a leader. However, both in the public discourse, and in the academic literature, the meaning of leadership has undergone substantial change during the timeframe of our sample.

\begin{figure}[h]
    \centering
    \efbox{\includegraphics[width=1\textwidth]{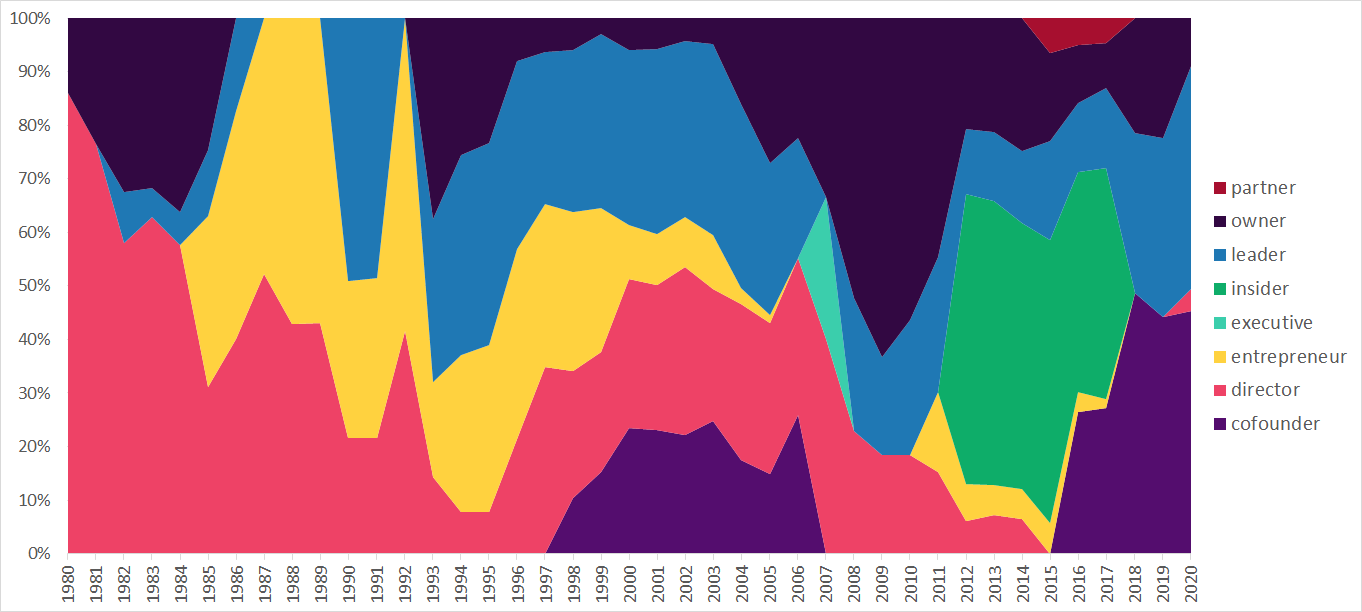}}
    \caption[Eight most semantically related role clusters to "founder" over time]{Eight most semantically related role clusters to "founder". Names are given by most word with strongest tie to "founder". Cluster for the word "CEO" is not shown.}
    \label{fig:c1:yoy}
\end{figure}

To account for such changes, we plot the semantic composition of "founder" over time in figure \ref{fig:c1:yoy}. That is, for each year in our sample, we calculate the relative weight of semantic similarity ties accruing to each of the clusters we previously identified.\footnote{We use a bi-directional 2-year moving average around the focal year, however, conclusions do not change when we use only the focal year.} Crucially, while the graph shows the weighted average of the semantic similarity for each cluster, we label each cluster with the  word that is closest to "founder". This allows the most prominent term to change, indicating a change in the meaning of the associated cluster.

Figure \ref{fig:c1:yoy} shows that the meaning of "founder" has changed significantly throughout our sample. In particular, our data now indicate that the overwhelming association of "founder" with administrative and hierarchical roles holds only insofar as we consider the semantic network generated across forty years of articles. Allowing these networks to change on a yearly basis identifies several distinct periods in which "founder" had a substantively different meaning in the language of  \textit{HBR}.

At the beginning of our sample, "founder" was semantically similar to "owner" and "director". From 1984 to 1993, however, the semantic cluster dominated by "entrepreneur" and "innovator" grew in similarity to "founder". During these years, founders described in  \textit{HBR}  were put in similar contexts and associated with similar behaviors as these creative roles. While the present article does not seek a causal explanation for the change in meaning of "founder", it should be noted that this period coincides with the end of a long recession and that several now well-known founders in the US technology sector, such as Steve Jobs, Bill Gates, and Michael Dell, rose to prominence during this time.

The second period, roughly ranging from 1990 to 2008 is characterized by the semantic cluster associated with leadership. Again, an informal explanation could assert that this period saw increasing interest in the definition of leadership roles and their distinction from other parts of the organization. For example, Kotter's influential article "What leaders really do" was published in  \textit{HBR}  in 1990. During this second period, the administrative / hierarchical clusters around "director" and "owner" continue to decline in similarity to founder. In the late nineties, however, hierarchical notions again gain prominence with the appearance of a semantic complex centered on "co-founder" and "patriarch". 

The years past 2010 are a final period in our sample. Much like the beginning period, it is characterized by a creative cluster around the words "insider", "outsider", and "newcomer". In contrast to that earlier period, there is no clear decline in the administrative and hierarchical role set. Instead, the role of a "founder" is remarkably varied, including a resurgence of the role of leadership and the semantic complex around "co-founder" (which we classify as administrative, but which has a strong relational component). Additionally, the idea of a founder as owner and proprietor is prevalent after 2008.

In conclusion, the meaning of "founder" has shifted over time. It was initially associated with administrative roles. We next identify a creative phase in the mid eighties, followed by a creative and social/organizational phase during the nineties and early two-thousands. Finally, towards the end of our sample, the meaning of "founder" becomes varied, including dominant phases of all our chosen categories.

\subsection{Contextual composition}
Our model of semantics is based on the core assumption that meaning is contextual. Thus, we next inquire which contexts, according to the writers of \textit{Harvard Business Review}, are associated with the role of "founder". Our analysis is based on the context network $Q$, derived from dyads $\rho$ and $\mu$ according to the measure $\underline{q}_\mu(\rho)$ discussed earlier. We first derive the context neighbors of "founder", defined as those words with which "founder" is likely to appear jointly in any sequence. Then, we repeat this procedure for those alter tokens and thus extract the same kind of relationships between any contextual terms associated with "founder".

Next, we again employ a hierarchical clustering algorithm. The groups of words we extract are terms that are likely to appear jointly throughout  \textit{HBR}. In a context network, the clustering algorithm can no longer disambiguate syntax and semantics. Instead, contextual clusters include all parts of speech. We denote these clusters by the word with the strongest contextual tie to "founder".

Our clustering algorithm again identifies several levels of contextual clusters. For clarity, we focus on the first level of clustering, which produces four large contexts relevant to the role of "founder", and two outlier groups that lack strong contextual ties to "founder". The four clusters seem to define different contexts in which a founder is active: in his or her \textbf{company}, as \textbf{CEO}, in a \textbf{market}, and with \textbf{people}. 
Table \ref{tbl:c1:context_cluster1} shows that "founder" is by far most likely to appear in sentences from the company cluster.
\begin{center}
\begin{table}[ht]
    \centering\small
 \begin{tabular}{||l | c||}
 \hline
 \hline
 Cluster &  Relative Weight \\ [0.5ex]
 \hline
\hline
 \multicolumn{1}{|l|}{\textbf{Company}} & \textbf{75$\%$} \\
\hline
\hline
company-insider-year-start-country
 &42$\%$ \\
\hline
engineering-industryadjusted-return-tie
 &33$\%$ \\
\hline
industry-change-capitalization-states-united
 &25$\%$ \\
\hline
\hline
 \multicolumn{1}{|l|}{\textbf{CEO}} &\textbf{ 11$\%$ }\\
\hline
\hline
 "CEO"-president-chairman-group-former &38$\%$ \\
\hline
steve-lewis-stories-walter-sergey
 &31$\%$ \\
\hline
new-comment-johnson-mckinsey-psychoanalytic
 &31$\%$ \\
\hline
\hline
 \multicolumn{1}{|l|}{\textbf{People}} &\textbf{ 7$\%$ }\\
\hline
\hline
top-executives-managers-members-boards
 &38$\%$ \\
\hline
people-best-employees-good-way
 &33$\%$ \\
\hline
organization-team-innovation-experience-leaders
 &30$\%$ \\
\hline
\hline
 \multicolumn{1}{|l|}{\textbf{Market}}  &\textbf{6$\%$} \\
\hline
\hline
market-companies-firm-large-small
 &41$\%$ \\
\hline
technology-services-including-amazon-ibm
 &30$\%$ \\
\hline
consumer-goods-high-instance-product
 &29$\%$ \\
\hline
\hline
\end{tabular}
    \caption[Contextual clusters for "founder"]{In bold, the four first level clusters act as categories, with weights as relative likelihood of occurrence within the context of "founder". For each category, three second-level sub-clusters most likely to occur in the context of "founder" are given. They are named by the five most proximate tokens.}
    \label{tbl:c1:context_cluster1}
\end{table}
\end{center}

\begin{landscape}
\begin{figure}
\begin{adjustbox}{minipage=0.45\linewidth,frame=0pt 1pt}
\includegraphics[width=0.98\linewidth]{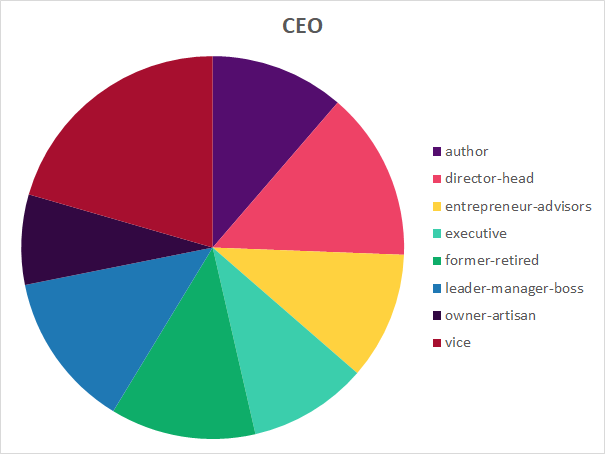}
\smallskip

\includegraphics[width=0.98\linewidth]{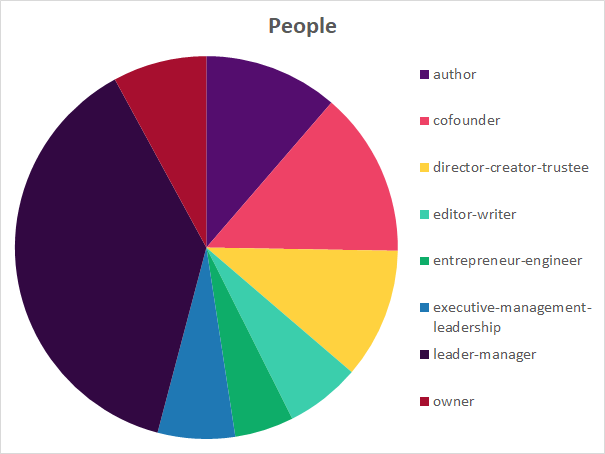}
\end{adjustbox}
\hfill
\begin{adjustbox}{minipage=0.45\linewidth,frame=0pt 1pt}
\includegraphics[width=0.98\linewidth]{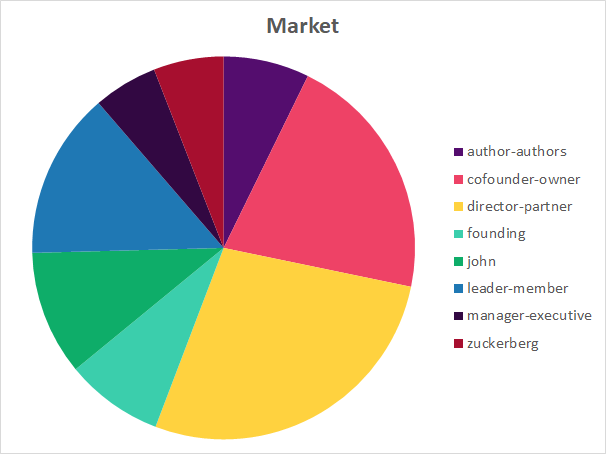}
\smallskip

\includegraphics[width=0.98\linewidth]{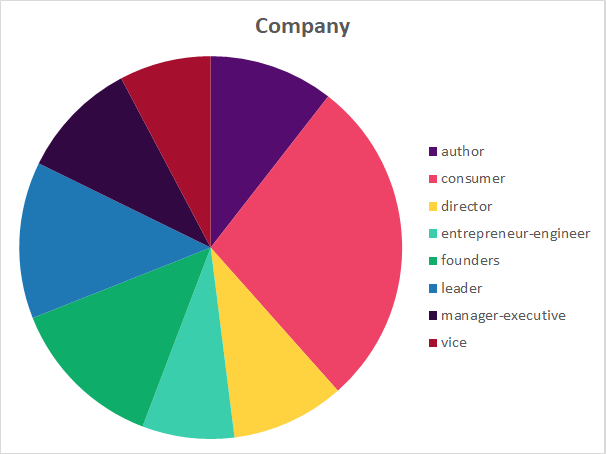}
\end{adjustbox}
\caption[Contextual roles of "founder"]{Eight most prominent semantic clusters of roles analogous to "founder" in the four defining contexts of "founder". Clusters are named by roles most analogous to "founder". For each context, ninety percent of the sum of tie weights are retained, and remaining low-valued ties are deleted.}\label{fig:c1:context_comp}
\end{figure}
\end{landscape}

To judge the variability of founder across these contexts, we analyze its substantive meaning for each contextual cluster. Figure \ref{fig:c1:context_comp} gives these compositions across all years.
The contexts of \textit{market}, \textit{people} and \textit{company} show roles associated with "founder". In the context of \textit{markets}, founders are similar to \textit{directors}, \textit{cofounders}, \textit{partners} or \textit{owners}. In the context of \textit{people}, on the other hand, a founder is overwhelmingly a "leader" or \textit{manager}. Interestingly, in the context of \textit{companies}, the defining term is "consumer". "Consumer", according to the language model, is not analogous to "founder" in other contexts. 
In contrast, in the context of "CEO", which includes discussions of the position, of people associated with the role of founder and prominent companies, founders appear in a variety of administrative and hierarchical roles.

Note also that while roles analogous to "founder" are similar to those identified in the overall analysis, cluster compositions have changed. For example, in the context of \textit{people} and \textit{company}, the terms "entrepreneur" and "engineer" define a role cluster, but in the context of "CEO", "entrepreneur" is similar to "advisors". Since clusters are defined by roles that are semantically similar, the results suggest that these roles have a different meaning in these contexts even as an analogy to founder.

In contexts addressing specific roles like "CEO", the concept of a founder is defined by analogous roles that are themselves partly ambiguous. Instead, a distinct meaning of "founder" seems to arise in contexts associated with markets, people, organizations, companies, engineering and so on. For that reason, we next analyze the roles of a founder by focusing on specific companies. 
\newpage
\newpage
\subsection{Founders of Google, Microsoft, Facebook and Apple}

As discussed in the prior sections, the identity of a founder is determined by a collection of roles that arise during different organizational contexts. In particular, the requirements of the role of a founder change drastically over the course of an organization's lifetime.

Four technology companies are especially prominent in the discourse of\textit{ Harvard Business Review}: Google, Microsoft, Facebook and Apple. We next condition the semantic network on sequences that are semantically related to any of the four firms. That is, we consider sentences in\textit{ Harvard Business Review} where, for example, the word "Google" has a high likelihood of appearing.
This contextual set includes sentences where the word "Google" does appear, as well as sentences that are situated in the semantic context of "Google", even if the term does not directly appear.

\begin{landscape}
\begin{figure}
\begin{adjustbox}{minipage=0.5\linewidth,frame=0pt 1pt}
\includegraphics[width=0.98\linewidth]{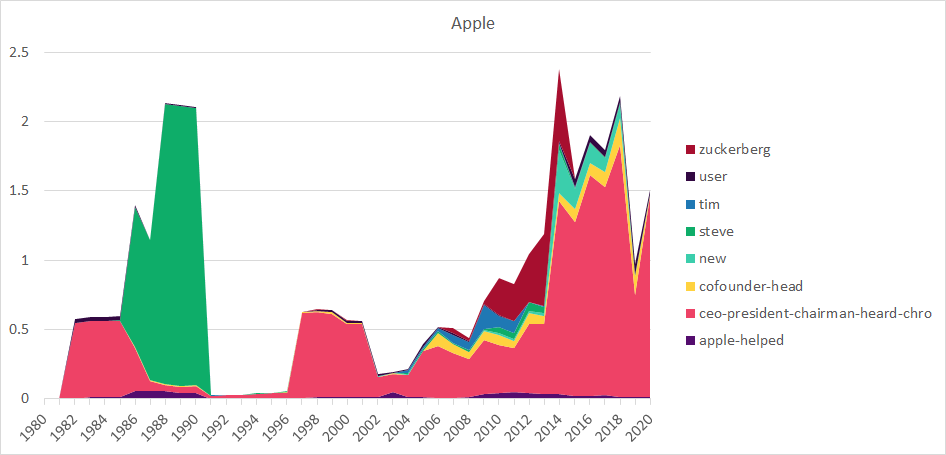}
\smallskip

\includegraphics[width=0.98\linewidth]{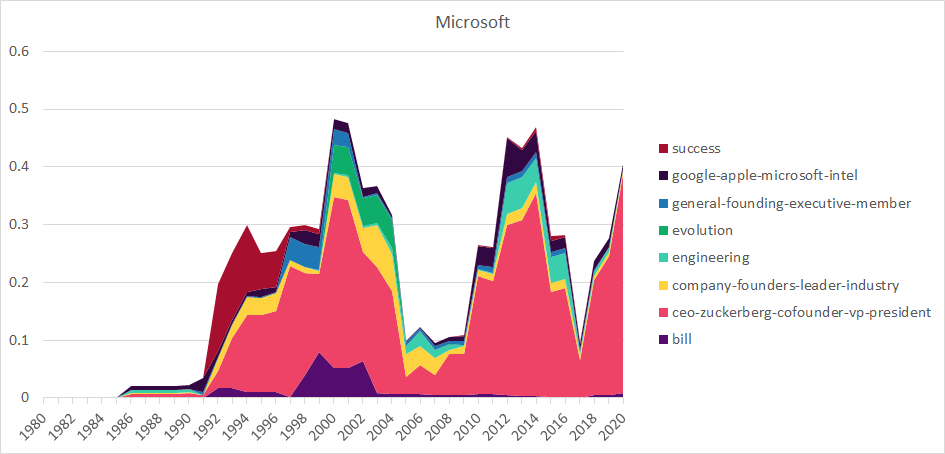}
\end{adjustbox}
\hfill
\begin{adjustbox}{minipage=0.5\linewidth,frame=0pt 1pt}
\includegraphics[width=0.98\linewidth]{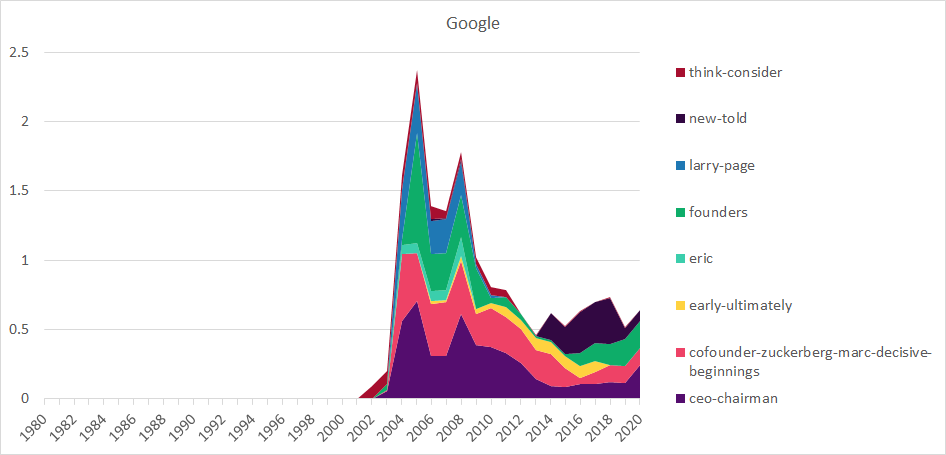}
\smallskip

\includegraphics[width=0.98\linewidth]{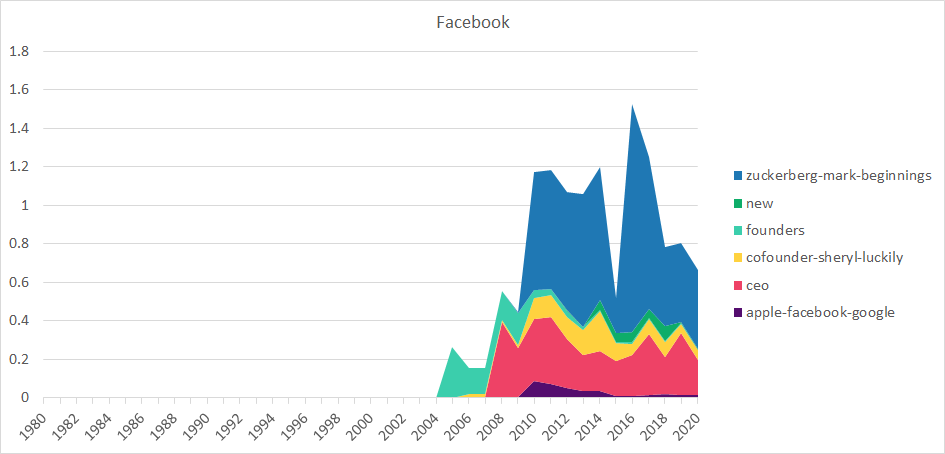}
\end{adjustbox}
\caption[Semantic clusters associated with "founder" in the context of four companies]{Semantic clusters associated with "founder" in the context of four companies. Each cluster is represented by the terms most semantically similar to "founder" in a given year.}\label{fig:c1:fourcompanies}
\end{figure}
\end{landscape}

In what follows we present the most semantically similar clusters to "founder" on a higher aggregation level, that is, without focusing specifically on roles.  Figure \ref{fig:c1:fourcompanies} depicts these four sets of semantic clusters over time. Note that instead of showing the composition of "founder" in terms of semantic clusters, we now show the sum of substitute tie weights for each cluster. This allows the comparison between the scale of semantic similarity and the degree to which the context of each company was significant to the meaning of founder over time.

Figure \ref{fig:c1:fourcompanies} suggests that our model tracks the quantity of discourse related to each company over time, including in sequences where the focal company does not directly appear. A role cluster around the word "CEO" is prominent in each context, however, its composition changes. Over the entire time horizon, the semantics of founder in the context of Apple and Microsoft are similar, while they differ for Facebook and Google.
Apple and Microsoft, companies that rose to prominence in the 1980's, display a stronger association between founder and singular administrative roles, such as \textit{president}. By contrast, both Facebook and Google initially seem to de-emphasize the singular "founder", in favor of the plural "founders" or "co-founders". 

Another regularity is visible when "founder" is increasingly associated with a person in the context of a focal firm. On average, this association develops between two and six years after the company became a relevant context in \textit{Harvard Business Review}. 
In particular, both Steve Jobs' and Bill Gates' contribution to the semantic identity of "founder" arises only after they left the position of "CEO" at their respective companies. Thus, while the concept of "founder" seems, for each company, tied to the prominence of a person, it is not necessarily tied to that person's function within the company.

We further observe that Mark Zuckerberg and Steve Jobs, from Facebook and Apple respectively, transmit the most meaning to "founder" from the set of names that appear in \textit{Harvard Business Review}. While Steve Jobs strongly defines the concept of "founder" for Apple during the 1980s, the magnitude of association for Mark Zuckerberg is less in any particular time period. On the other hand, Mark Zuckerberg also appears as related term when authors of \textit{Harvard Business Review} refer to firms other than Facebook. While clusters are fixed over time in figure \ref{fig:c1:fourcompanies}, further analyses show that Mark Zuckerberg is a defining element of the identity of "founder" in the context of each of the four firms starting in 2010 up until around 2015. It is further notable that Facebook's strong association with its founder leads to less top-level clusters compared to the other companies.

Finally, "founder" in the context of Microsoft displays a trajectory of associated terms, starting with "success" during the early and mid nineties, moving towards "evolution" during the early two thousands and circling back toward "engineering" after 2010. For Apple, minor associated clusters center around "consumers" and the term "new".
\newpage
\newpage
\subsection{Semantic importance of "founder"}
\begin{figure}[h]
    \centering
    \efbox{\includegraphics[width=0.7\textwidth]{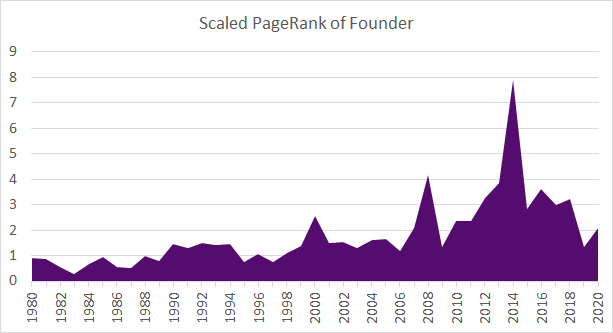}}
    \caption[Scaled PageRank centrality of "founder" for a given year]{Scaled PageRank centrality of "founder" across the entirety of \textit{Harvard Business Review} for a given year.}
    \label{fig:c1:pr-founder}
\end{figure}

In the final part of our descriptive analysis, we take a step beyond the immediate semantic ties of the word "founder". Instead, we now lay focus on the entire semantic substitution network and analyze the semantic importance of "founder" and its analogous roles, as measured by the PageRank of a term in the semantic network. Figure \ref{fig:c1:pr-founder} shows that the semantic importance of "founder" increased, peaking around 2013 to 2014. This development is consistent with the importance of new technology companies, such as Facebook, and their founders, such as Mark Zuckerberg.
\efboxsetup{linecolor=black,linewidth=1pt}
The role of a founder seems more central in defining meaning, in part due to its association with important individuals, and its increasing association with the role of CEO. Both "founder" and "CEO" share a similar trajectory in semantic importance.

\begin{figure}[h]
    \centering
    \efbox{\includegraphics[width=0.9\textwidth]{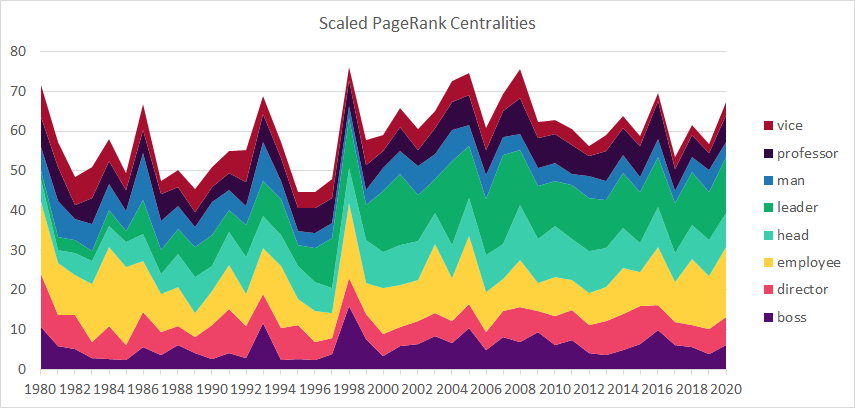}}
    \caption[Highest PageRank centralities of roles analogous to founder across]{The eight highest PageRank centralities of roles analogous to "founder" across the entirety of Harvard Business Review for a given year. "CEO" omitted.}
    \label{fig:c1:pr-others}
\end{figure}

\begin{figure}[h]
    \centering
    \efbox{\includegraphics[width=0.9\textwidth]{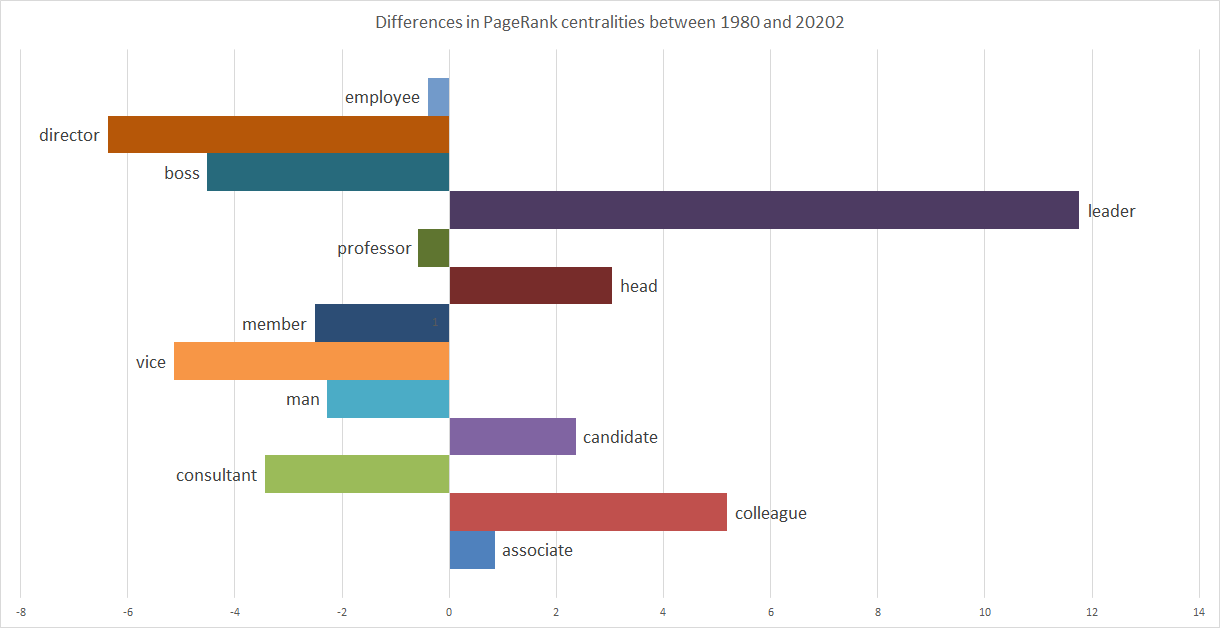}}
    \caption[The largest changes in PageRank centralities of roles analogous to "founder"]{The largest changes in PageRank centralities of roles analogous to "founder" across the entirety of \textit{Harvard Business Review} from 1980 to 2020.}
    \label{fig:c1:pg_delta}
\end{figure}

Next, we also check the trajectories of the other most semantically important words that are associated with "founder". Figure \ref{fig:c1:pr-others} indicates that most of these terms did not exhibit systematic changes that could explain the trajectory of "founder". The results therefore suggest that, compared to other roles, "founder" comes to define more of the meaning in \textit{Harvard Business Review}, especially after 2012.
Nevertheless, Figure \ref{fig:c1:pg_delta} suggests that between 1980 and 2020, some roles have seen significant changes in semantic importance. Most prominent would be the rise of the role of "leader". The decreasing semantic importance of "man" as component of central roles in \textit{Harvard Business Review} is a positive note.

\section{Conclusion}

The present work proposes a novel method to extract contextual semantic substitution networks from deep language models, in particular transformer architectures such as BERT. The considerable advances these models offer in the representation, generation, and understanding of natural language indicate their potential for the analysis of existing texts and, in particular, for the analysis of lexical semantics. However, a direct application of their representations has proven difficult. To the best of our knowledge, the present work is the first to derive contextual semantic substitution networks from the outputs of the language model. We propose several benefits of this approach: the model does not require scaling or aligning across contexts and trained instances, has a formal representation without free parameters, is congruent with the training task of the language model, incurs no loss of information relative to the language model, can identify and condition on arbitrary context sets, allows for inference on any level from global- to occurrence level, and allows the use of a wide array of tools from network analysis to gain insights from both local and global properties of language.

As an empirical demonstration, we perform a descriptive semantic analysis of the term “founder” in the entire corpus of \textit{Harvard Business Review} from 1980 to 2020. Instead of extracting word-senses of the relatively few direct occurrences of “founder” in the text, we query the language model for its representation of the semantics of “founder”. We find, in contrast to much of the literature, that the semantic position of “founder” in \textit{Harvard Business Review} is equally defined by administrative and creative roles. Analyzing this relationship over different years, we find that the meaning of “founder” has changed considerably. While a dominant association with administrative roles exists early in our sample, we also identify creative and entrepreneurial phases, phases with focus on social and organizational roles and, in the last part of our sample, a conception of “founder” that is multifaceted and focused on collaboration and leadership. Next, instead of defining contexts beforehand, we use our method to identify and cluster contexts in which the word “founder” may appear. Here, we find four large semantic complexes in which “founder” appears: in a market, in the company, with people, and as CEO. We then analyze the meaning of “founder” within these contexts and find that “founder”, when invoked in the context of “CEO”, is defined by its positions in a company hierarchy. However, such analogies with company positions do not clarify the meaning of the role of founder.  The semantics of founders are less ambiguous when they lead and manage in contexts associated with people or organizations, or when they focus on consumers, or when they appear in sentences discussing companies, industries, or engineering efforts. Overall, the semantics of “founder” are difficult to nail down because founders have a multitude of roles, with different semantic associations across contexts. We conclude that a semantic analysis of “founder” needs to be highly contextual, and, as illustration, use our method to analyze the context of four prominent technology companies separately. Here, we find that the concept of a founder changes both across companies and over time. For two established firms, the role of founder has a strong association with the role of the CEO and the individual who created the company. The latter, however, arises after that person has left the position of CEO, such that we suspect that these roles are substitutes. The two newer technology companies convey a different concept of who their founders are. The differences are both in the association with central individuals, and in the roles attributed to them. In either case, the semantic trajectory of “founder” differs from that of established firms. These recent changes, we find, have increased the semantic centrality of “founder” in the corpus of \textit{Harvard Business Review}, a change that is not driven by the importance of other roles analogous to founder. We therefore conclude that founders have become increasingly important in defining the meaning of business-related texts published in the past decade.

The present analysis, we believe, only scratches the surface of what can be accomplished with contextual semantic networks. Indeed, the results derived in this work are primarily based on standard measures of relatedness for a focal term in broadly-construed sets of contexts. Nevertheless, structural and contextual measures of semantics allow for a more general and less categorical understanding of meaning. Thus, in addition to running such analyses for further terms of interests and in different corpora, our method may also shed new light on issues of gender, political, and linguistic polarization, and on the transmission and propagation of linguistic change in different arenas. Furthermore, our method also opens new avenues in computational linguistics, most of which we leave for future research. First, the directed semantic substitution networks allow for the distinction between semantics that are received and conferred in a given context. Second, the structure of contextual semantic substitution networks may allow for the identification and generalization of different types of semantic relations; this includes antonymy and hyper- or hyponymy, but could also include higher-order labels traditionally associated with semantic networks. For example, combining contextual and synonymy networks could identify verbs, adjectives, or locations associated with a focal term. Third, our focus on single words is restrictive. Instead, semantic networks call for analyses that define and track semantic concepts, prototypes and categories. Here, the large set of network analysis tools allows for the generation of new insights into language use, by going beyond local relations of a word.

 Finally, we see the combination of contextual semantic networks with other data and statistical methods as especially promising. On the one hand, this may include sequential and ontological data, e.g., in the analysis of discourse between more than one author or speaker. On the other hand, the use of exogenous variation may allow for causal analyses of structure and dynamics of semantics networks of language use.
 
Although not the focus of the present work, we finally note that semantic substitution networks can be employed more generally to analyze the behavior of deep learning models. The formal framework establishes a probabilistic interpretation of a transformer network. The weighted and directed networks, and their associated measures such as entropy, represent the conditional dependence structure of the underlying model as it responds to input vectors. Such a graphical representation of a transformer model could inform research on these model architectures in numerous ways. For example, a network based on entropy could be combined with substitute relations to measure behavior of the transformer architecture not only for specific sequences, but for specific relations between input tokens. Not only could such methods provide more structured insights into the performance of these models, they could also allow the researcher to separate behavior from the underlying data and trained instance.  We leave these avenues to further research.

\newpage
\appendix

\section{Appendix}

\subsection{Notation}\label{app:ch1:notation}
The set $\Omega$ consists of atomic elements we call \textit{tokens} word \textit{words}. These tokens will form the nodes of our network and we denote them with symbols such as $\mu$ and $\tau$. In our example, these are vocabulary words such as \textit{dog} or \textit{cat}.

The data consists of a set $D$ of sequences\footnote{Our data includes $|D|=T$ sequences. If we wish to refer to a specific sequence in that data, we could write $s_t=(s_{1,t},s_{2,t},\ldots,s_{i,t}, \ldots, s_{m,t})$. For what follows, distinguishing $t$ is not necessary, as predictions are fully determined by a given sequence. We also suppress the explicit mention of the condition that $s \in D$.} of the form $s=(s_{1},s_{2},\ldots,s_{i}, \ldots, s_{m})$ with $s_{i} = \mu \in \Omega$. In our example, such sequences are sentences, for example $s= (A, cat, is, a, pet,.)$. We write $s$ to denote any such sequence, whether it occurs in our data $D$ or not.
For a given $s_{i}$, we denote the remaining elements of the sequence as $s_{-i}=(s_{1},s_{2},\ldots,s_{i-1},s_{i+1},\ldots)$.

In what follows, we condition on certain subsets of $D$, such as sentences written in a given year or sentences that include the word \textit{pet}. We denote an such an arbitrary subset $C \subseteq D$ as a \textit{context}.

Given the observed sequence $s$, the language model $P_D$ generates predictions for any element $s_{i}$ of a sequence $s$. This prediction is based on a (conditional) probability measure. 

Whereas $s$ is a particular sequence, usually one from the focal corpus, $w$ is its probabilistic representation. That is, we write $w_{i}$ to denote the categorical probability distribution the $i$-th position in $w$.\footnote{$w_{i}$ is short-hand for the event that position $i$ is a certain word \textit{in a given sequence}. That is, $w_i$ could be defined as a random variable pointing from this event to a integer for each word. More formally, we define $w_i$ as a random element from a measure space over the sequences $s \in D$ to $(\Omega, 2^{\Omega})$.} We distinguish $w$ and $s$. That is, $w_{i}$ is a random element, the observation of \textit{some} element is denoted as $s_{i}$, and we write $s_i=\tau$ if we have observed some \textit{specific} word $\tau$. 

Since we always observe some $s_i$, $w_i$ is counterfactual from the perspective of the researcher. The event $\{ w_i=\mu, \textit{if}~ s_i=\tau\}$ can thus be interpreted as $\mu$ substituting for $\tau$ at position $i$ in $w=s$. Similarly, a substitution at any position in $s$ is the event $\{w_{\tau}=\mu\}$ = $\cup_i\{ w_i=\mu, s_i=\tau\}$. Note that for the model, $w_i$ is not counterfactual, as $s_i$ is not provided.

To simplify notation and to indicate knowledge of $s_i$, we add a reference to the conditioning operation
$$ P( \{ w_i=\mu,  s_i=\tau\}|w_{-i}=s_{-i}) \defeq P(w_i| s_i, w_{-i}=s_{-i}) \defeq P_D\left(w_{i}=\mu| s_i, s_{-i}\right) = P(w_i|s)
$$
that is, we simplify notation as $P(.|s)$ and we assume the elements of $w$ and $s$ are allocated as realizations of random variables, and as information set of the researcher, in a way that is consistent with our model, as will be discussed below.

\subsection{Structural language models}\label{app:ch1:scope}

We seek a computational representation of language that is powerful enough to infer information about words in a sequence, as opposed to words in the entirety of a corpus. Such a model must encode a relationship between a word $\mu$  at position $i$, and other elements in a sentence $w$,  that is, a \textit{contextual} model $P(w_i = \mu| w_{-i})$ in $D$.

One may then ask whether such a representation can be estimated without imposing a model. We might choose to use any sort of count or frequency information, for example, to estimate 
$$\frac{P(w_i = \mu, w_{-i})}{P(w_{-i})}=P(w_i = \mu| w_{-i})$$

However, lacking a model, we can not aggregate across \textit{different} sequences to estimate the join probability $P(w_i = \mu, w_{-i})$. The majority of sequences are unique, such that for any given context $C\subseteq D$ we have that $P(w_i = \mu, w_{-i}) \approx \frac{1}{|C|}$.
A similar argument holds for the informational content of  $P(w_{-i})$. For each $w_{-i}$, we require several observations of the exact same $w_{-i}$ with  differing observations of $w_i$.

Since every token in the dictionary is ex-ante distinct, and since sequences can easily have ten, twenty or even more tokens, such estimations suffer from the curse of dimensionality: Even big amounts of text are sparse compared to all sequences that \textit{could have been written}, and as $D$ grows, further observations increase the need to estimate for new sequences $w_{-i}$ in excess of providing information about existing realizations of $w_{-i}$. The set of $w_{-i}$ for which multiple observations with differing $w_i$ are small, and the set of sequences $w$ for which multiple observations exist for every position $i=1,\ldots$ goes to the empty set. Hence, it is practically impossible to construct $P(w_i=\mu|w_{-i})$ with simple statistics.

Instead, our estimation needs to relate the information gained  from one possible $w^1_{-i}$ to that of another, distinct $w^2_{-i}$. Only then can we estimate its impact on the likelihood of $w_i$. In consequence, to define $P(w_i=\mu|w_{-i})$ for every $w_{-i}$ we might care about, our procedure must be able to relate the constituent parts of $w_{-i}$ to each other. Thus, we must specify the relationships within the distribution of $(w_i, w_{-i})$ for every position $i$ for a single sequence, rather than for a moment or aggregate statistics. This means that the procedure is inferring counterfactual outcomes and our statistical understanding of language in the above sense will be based on a \textit{structural} model.

Finally, the corpus $D$ is a product of the author's style, his or her intention for writing a particular sentence, the audience the text is written towards, and the editing process it has to go through to see the light of day. That is, the generation of language includes factors such as knowledge and intent, that are not visible from the text itself. Similarly, the reception of language involves an agent imbued with knowledge, tastes and other properties that remain latent from the writings our algorithm operates on (\cite{nirenburg2004ontological}). Due to this distinction between an extant collection of writings and their underlying cause, \cite{bender_climbing_2020} question whether computational models capture real meaning, or whether they instead merely imitate the form of communication. 

The present work takes a pragmatic stance on the issue.
First, the relations estimated by the language model characterize the use of language within the given sample. Second, as detailed in the following two sections, the unobserved factors mentioned in the prior paragraph appear in the form of latent distributions in our probabilistic framework. These distributions can be understood as causal mechanisms that determine which sequences are ultimately written. We propose to estimate these distributions from the corpus, which serves to define an representation of language \textit{internal} to these texts. Inferences thus remain a matter of identification, and as such depend in equal measure on the properties of the sample and assumptions about probability distributions that necessarily remain hidden to the observer and the model. Since these considerations are not the focus of the method we develop here, we speak of language as those linguistic relations and identities that appear in the corpus of interest, and leave further claims of cause and generality to the application.

\subsection{Semantic Substitution}\label{app:ch1:tie_overview}

\subsubsection{Model and assumptions}\label{app:ch1:assumptions}

We derive all our measures from the language model's conditional probability estimates of the form
$$P_D\left(w_{i} | w_{-i}=(s_{1},\ldots,s_{i-1},s_{i+1},\ldots)\right)$$
or simply $P\left(w_{i}| w_{-i}=s_{-i}\right)$ for the observed (or hypothetically given) sequence $s=(s_i,s_{-i})$.

$s_{-i}$ fully specifies the relevant inputs for the model $P_D$, that is,  $P_D\left(w_{i} | w_{-i}=s_{-i}\right) = P_D\left(w_{i} |  w_{-i}=s_{-i}, \alpha \right)$ for any information $\alpha$. For example, BERT does not use $s_i$ to predict $w_i$, and so $P_D\left(w_{i}| w_{-i}=s_{-i}\right)=P_D\left(w_{i}|w_{-i}=s_{-i},s_i\right)$. We assume this specification characterizes the underlying data generating process:
\paragraph{\textbf{Assumption: Counterfactual assumption.}}
\begin{align}\label{app:ch1:ass1}
P\left(w_i | \text{Input} \right)  = P\left(w_i | s_{-i}, s_i \right)= P\left(w_i | s\right)
\end{align}
where in our case $P\left(w_i | \text{Input} \right) = P\left(w_i | s_{-i}=w_{-i}\right)$. That is, even if the data provides instances of $(s_i=\mu, s_{-i})$ and $(s_i=\tau, s_{-i})$ with $\mu \neq \tau$, these are draws from the same random variable $w_{i}$, which is uniquely identified for a given $s_{-i}$.
The assumption is consistent with the output of most transformer models, including applications of dynamic patterns as in \cite{amramiBetterSubstitutionbasedWord2019}. If, however, the input is more informative than the sequence under consideration, our derivations would need to be revised accordingly.

We further assume that model outputs are independent conditional on an input sequence. That is
\paragraph{\textbf{Assumption: Contextual independence.}}
\begin{align}\label{app:ch1:ass2}
P\left(w_i, w_j | s \right)  = P\left(w_i | s\right)  P\left(w_j | s\right)
\end{align}
Again, this assumption seems to be trivial if we query BERT for single positions in a given sequence. In particular, since BERT makes use of all context to arrive at a prediction of $w_i$, it is fully defined by that context and $P_D\left(w_i | s\right) = P_D\left(w_i | w_j, s\right)$. Nevertheless, the assumption is not trivially true. Note that under assumption \eqref{app:ch1:ass1}, assumption \eqref{app:ch1:ass2} implies that 
$P\left(w_i, w_j | s \right) =  P\left(w_i | s_{-i}\right) P\left(w_j | s_{-j}\right)$. The reader may note that this is not the only way to derive the given quantity from BERT. For example, BERT also outputs $P_D\left(w_i, w_j | s_{-i,j} \right) \neq P_D\left(w_i | s_{-i}\right) P_D\left(w_j | s_{-j}\right)$.
However, for a given sequence $s$, the  output $P_D\left(w_i, w_j | s_{-i,j} \right)$ is clearly less informative than conditioning on the full sequence. Since our objective is to analyze existing text, and make maximal use of information, we query BERT for single positions, giving us estimates that are conditionally independent.

For expositional purposes we may assume that each token occurs as most once in each sentence: $\forall i: s_{i}=\tau \Rightarrow \tau \notin s_{-i}$. This assumption simplifies the formal derivations when focusing on sequences, but we will indicate when it is used. Given assumption \eqref{app:ch1:ass2}, all  results in question can also be derived for occurrences instead of sequences.

\subsubsection{Approach and input}

Consider a given sequence $s$ in which token $\tau$ occurs, and another arbitrary token $\mu$ from $\Omega$. Then the value
$$P_D\left(w_{i}=\mu |w_{-i}=s_{-i}, s_i=\tau\right) \eqdef P_D\left(w_{i}=\mu| s_i=\tau, s_{-i}\right)$$
was defined as the \textit{conditional probability of $\mu$ replacing $\tau$} in sequence $s$. In that sense, $\mu$ is the focal token, whereas we condition on $\tau$ as element of the input sequence. Recall that we add the reference to $s_i$ to clarify that token $\tau$ at position $s_i$ is the object of our analysis. That is, we use assumption \eqref{app:ch1:ass1}.

The above probability implies a necessary, but not sufficient condition for either semantic or syntactic substitutability. A zero probability indicates which words \textit{cannot} replace each other: A word is not a substitute, when it is not appropriate in terms of syntax and semantics. The condition is jointly sufficient for syntax, semantics and other indicators that the model learns from data. Research indicates (\cite{laicherExplainingImprovingBERT2021}) that for contextual embeddings, orthographic similarity is another condition. However, we do not observe such orthographic relations while employing our network approach.

In what follows, we will derive two notions of substitution of a focal token: \textit{aggregate} and \textit{compositional}, which represent two different ways of aggregating across sequences.
Before formally defining both \textit{aggregate} and \textit{compositional} ties, we need to understand the relationship between measures across multiple sequences.

\subsubsection{Aggregation across sentences}\label{app:ch1:aggregation}
To make useful statements about the text in question, we need to aggregate across sets $C$ conditioning $P$. That is, lifting our measures from specific sequences towards a larger context, say a period of time, requires a sense of what sequences $s$ are to be analyzed. As mentioned in \ref{app:ch1:scope}, which sentences \textit{appear} depends on the senders, the recipients and ultimately the intents of communication. For every sentence we observe, uncountable others were conceivable and yet, were never written.

Thus, statistical identification lacks the true \textit{prior distribution of sentences} in a given context: the probability $P(w=s)$ and in particular $P(w=s|\tau \in s)$ of a sequence $s$ when a focal token $\tau$ is observed. To weigh our information of $\tau$'s likely substitutions, we need to know in what sentences $s$ it is likely to be used.

The language model we consider could generate substitution probabilities for any sequence $s$, whether it is part of the original data set $D$ or not. Thus, if substitution relations were generated from an exhaustive collection of sentences $s$, these relations and the model $P$ would be equivalent. 
If $P(w_{\tau}|\tau \in s)$ were known, any required sentence could be analyzed and, as shown below, measures could be correctly aggregated across contexts.
However, any true $P(w_{\tau}|\tau \in s)$ remains latent. Thus, our definition of $P(w_{\tau}|\tau \in s)$ fundamentally defines \textit{the object of analysis}. 

To start, the empirical distribution of sequences in $D$ provides an estimate of $P(w=s|\tau \in s)$. In accordance to our main objective, our object of analysis is then the \textit{language use in a given corpus}.

Assume first that $\tau$ can occur at most once in any sequence in $D$. We then estimate $P(w_{\tau}|\tau \in s)$ from data as
$P_D(w_{\tau}|\tau \in s)=\frac{1}{|\{\tau \in s\}|}$. We can further condition this probability on some context $C$ as $P_D(s|\tau \in s, s \in C)=\frac{1}{|\{\tau \in s, s\in C\}|}$. 

To shorten notation, we will denote the event that $\tau \in s, s \in C$ as $\tau \in C$ if doing so leads to no ambiguity. The estimate for a given $C$ becomes
$$P_D(w=s|\tau \in s, s \in C)=\frac{1}{|\{\tau \in C\}|}$$

Assume now that that $\tau$ can occur any number of times in a given sequence. We must then consider the event $\tau \in s=\cup_i \{s_i = \tau, s_{-i}\}$ such that we can estimate the above quantity instead as
$$
P_D(w=s|\tau \in s, s \in C)=\frac{|\cup_i \{s_i = \tau, s_{-i}\}|}{|\{\tau \in C\}|}
$$
Note in particular, that if $\tau$ occurs $k$ times in $s$, we have that $P_D(w=s|\tau \in s, s \in C)=\frac{k}{|\{\tau \in C\}|}=\sum_k\frac{1}{|\{\tau \in C\}|}$. We use $P_D$ to indicate that these probabilities will be extracted from the language model.\footnote{That is, $|\{\tau \in s, s\in C\}|$ is given as $\sum_{\tau \in s \in C} \sum_{\mu} g_{\mu \tau}(s)$, and $g$ will be defined below.} 

The need to estimate the distribution of sequences illustrates the dependence of our result on the corpus of interest.
For computational purposes, it will sometimes be useful to focus on occurrences rather than distributions of sequences. Several sequences in the corpus may be identical, say $s^{(1)}=s^{(2)}=s^{(3)}$, however, this leads to no structural differences in how the language model predicts focal tokens. If a sequence occurs $n$ times, we derive $n$ separate, but identical sets of predictions. 
Hence, in our computational approach, when we sum over such occurrences, we sum over results related to $s^{(1)}, s^{(2)}$ and $s^{(3)}$. To indicate this we write
$$
P\left( s_i=\tau,s_{-i}| \tau \in s, s \in C\right) = \frac{1}{|\{\tau \in C\}|}
$$
using $P$ instead of $P_D$ to indicate that these values are not extracted from the language model.

In other words, if $\tau$ occurs more than once in a sequence, we can, without loss of generality, consider these occurrences to be unique in separate sequences. This fact will be useful to deal with positional information.

\subsubsection{Conditioning and substitution}

Both \textit{aggregate} and \textit{compositional} substitution are defined for any context $C$, including a single sequence $s$. However, the measures only differ across contexts larger than a single sequence. For that reason, we define both measures for a context $C$, letting a single sequence $s$ arise as a special case.

\subsubsection{Compositional Substitution}
Consider again the output of the language model
$$P_D\left(w_{i}=\mu | s_i=\tau,s_{-i}\right)$$
Recall that we write
$w_{\tau}= \cup_i \{w_i, s_i=\tau\}$  for a substitution at any position in $s$, and that we shorten $\tau \in s, s \in C$ to $\tau \in C$.
Marginalizing over $s$ gives the conditional distribution of substitution as a function of $C$:
\begin{align}
    \label{eqn:c1:FRagg:l1} P_D\left(w_{i} |  s_i=\tau,s_{-i}, s \in C\right) &\rightarrow P_D\left(w_{\tau} | s, \tau \in s, s \in C\right) \\
    \label{eqn:c1:FRagg:l2} \Rightarrow  P_D\left(w_{\tau} | \tau \in C\right) &= \sum_s P_D\left(w_{\tau} | s, \tau \in s, s \in C\right) P_D\left(s|\tau \in s, s \in C\right)
\end{align}
We make use of the assumption that $\{s_i=\tau,s_{-i}\}=\{\tau \in s\}$, such that the arrow in \eqref{eqn:c1:FRagg:l1}  is an equality. 
Assume instead that $\tau$ occurs $k\geq2$ times in $s$. Then, \eqref{eqn:c1:FRagg:l1} obviously underestimates the true value,  as we attempt to aggregate across positional information. However,  we can instead marginalize over each position, and hence each occurrence of $\tau$ in \eqref{eqn:c1:FRagg:l2}. $w_i$ is  uniquely defined in $P_D$ for a given occurrence $s_{-i}$
$$
P_D\left(w_{\tau} | \tau \in C\right) = \sum_{\tau \in s, s_i=\tau} P_D\left(w_i |  s_i=\tau,s_{-i}, s \in C\right) P\left( s_i=\tau,s_{-i}| \tau \in s, s \in C\right)
$$
however, $P\left( s_i=\tau,s_{-i}| \tau \in s, s \in C\right)$ is simply $1/|\{\tau \in C\}|$: Each occurrence of $\tau$ belongs to a unique sequence $s$ in $C$, and given our prior assumptions and the setup of BERT, multiple occurrences in the same sequence are treated independently. Alternatively, we can think of marginalizing over every $i \in s$ for the context $C$ for which $s_i=\tau$. The above formulation is set up this way: the event $\{\tau \in C\}$ encodes a number of occurrences rather than a number of sequences. In practice, we implement the second approach, but note that the differences are usually minor.

Crucially, we represent the relation between $\mu$ and $\tau$ by an edge in a multigraph $G$, with parallel edges indexed by sequences $s$. In the example above, the conditional probability implies a \textit{compositional substitution tie} $g_{\mu}(s,\tau)  \defeq P_D(w=\mu|\tau \in s)$ where the direction of an edge marks an instance in which $\mu$ replaces $\tau$.

$g_{\mu}(s,\tau)$ is the degree to which \textit{$\mu$ replaces $\tau$, given that $\tau$ occurs in $s$}. 
Note that the measure is conditional on both the occurrence of $\tau$ and the context in which it occurs. $\tau$ may be used in different ways and for different reasons throughout the corpus. However, a high value of $g_{\mu}(s,\tau)$ implies that in this instance, $\mu$ would serve a similar linguistic function and carry a similar meaning.

The support of $w$ can be specified as the set $\Omega$ with or without $\tau$ itself. If $\tau$ is included, reflexive values like $g_{\tau}(s,\tau)$ exist and indicate, for example, the confidence with which the language model sees $\tau$ as correct prediction if it occurs in $s$.\footnote{The model outputs $P_D\left(w_{i}=\tau | s_i=\tau,s_{-i}\right)$ given that $s_i$ remains unobserved in training and inference. This value is relevant when calculating entropy measures. To define \textit{substitution}, we naturally only consider the probability mass attributed to tokens $w_{i} \neq \tau$.} 

In general, $g_{\mu}(s,\tau) \neq g_{\tau}(s,\mu)$.
If $\tau$ occurs once in $s$, then we have that $\sum_{\mu} g_{\mu}(s,\tau)=1$. Conversely, $g_{\tau}(s,\mu)$ is the degree to which $\tau$ can replace another token $\mu$ that appears in $s$. Its sum $\sum_{\mu} g_{\tau}(s,\mu)$ can be larger than unity, if $\tau$ is a likely substitution for many words in $s$, or smaller than unity, if it is not.

Although we focus on aggregate substitution in our main analysis (see also next section), we note that compositional substitution can similarly be aggregated across contexts. Hence, we can also extract a simple graph for context $C$ from the multigraph $G$ with ties given as $P_D\left(w=\mu | \tau \in C\right)$.

Deriving $P_D\left(w=\mu | \tau \in C\right)$ for a given pair $\mu,\tau$ can be accomplished efficiently within our network $G$: We take a subgraph corresponding to all edges $s$ in the context $C$,\footnote{The context edges we introduce below allow for such queries without specifying all $s$ in that occur in $C$ separately.} and aggregate in-edges within nodes $\tau$ as
$$g_{\mu}(C,\tau) \coloneqq \sum_{\tau \in C} \frac{g_{\mu}(s,\tau) }{|\tau \in C|}$$
The transformed graph, denoted by \textit{compositional substitution graph}, is a simple directed graph specifying (compositional) substitution relations \textit{from $\mu$ towards $\tau$} for the context $C$, while keeping the probabilistic interpretation intact such that $g_{\mu}(C,\tau)=P_D\left(w=\mu| \tau \in C \right)$. 

The out-neighborhood of $\mu$ are those terms that $\mu$ can replace, and in general $\sum_{\tau} g_{\mu}(C,\tau) \neq 1$. If we refer to the compositional substitution ego-network or the compositional substitution neighborhood of $\mu$, we mean the out-neighborhood. Consequently, the compositional substitution ego-network for $\mu \in C$ requires running the language model over all $\tau \in s \in C$.

Conversely, the in-neighborhood of $\tau$ are terms that replace it, and we have\footnote{Since \begin{align*}
    \sum_{\mu} g_{\mu}(C,\tau)&=\sum_{\mu} P_D\left(w=\mu| \tau \in C \right) = \sum_{\mu}\sum_{s} P_D\left(w=\mu | s, \tau \in s, C\right) P_D\left(s|\tau \in s, C\right)\\ &= \sum_{s} P_D\left(s|\tau \in s, C\right) \sum_{\mu} P_D\left(w=\mu | s, \tau \in s, C\right) = \frac{1}{|\{\tau \in C\}|}\sum_{s \in C} \mathbbm{1}_{ \{\tau \in s\} } = 1
\end{align*}}  that $\sum_{\mu} g_{\mu}(C,\tau) = 1$. Deriving the in-neighborhood of $\tau$ requires running inference on its occurrences in $C$ only.

\subsubsection{Aggregate Substitution}

\textit{Compositional substitution} $g_{\mu}(s,\tau)$ is a function of both $s$ and $\tau$: It measures the likelihood of a substitution tie, but given the event that $\tau$ has occurred. As such, the probability of $\tau$ actually occurring in context $C$ plays no role in this measure.

Conversely, we can also derive the substitution characteristics of $\mu$ across the distribution of $\tau$ that are likely to occur in $C$.
That is, we now account for the likelihood that $\tau$ appears in $s$. 
In this case, we still derive a dyadic measure of association between two tokens, however that measure is no longer conditional on the occurrence of the alter token $\tau$.

As detailed in \ref{app:ch1:structural_identification}, our experiments show that this measure draws sharper distinctions between semantic and syntactic relations and allow us to identify part-of-speech clusters. These clusters do not arise with $g_{\mu}(s,\tau)$, since the measure of \textit{compositional substitution} is independent of the frequency of $\tau$ in the corpus.

Define a simple graph $G^C$ with ties between $\mu$ and $\tau$ given by
$$g_{\mu,\tau}(C) \defeq  P_D\left(w_{\tau}=\mu|s \in C\right) =  P_D\left(\cup_i \{w_i =\mu, s_i=\tau\}|s \in C\right) $$
as noted above, this measure is conditioned on the event $\{s \in C \}$, whereas \textit{compositional substitution}, which we derived as $P_D\left(w_{\tau} | \tau \in s, s \in C\right)=P_D\left(w_{\tau} | \tau \in C\right)$ is conditioned on $\{\tau \in s, s \in C\}$.

For a single sequence $s$, these measures are equal: $g_{\mu}(s,\tau)=g_{\mu,\tau}(s)$, since $\tau$ either occurs or it does not. 
Across a wider set of contexts, we can derive
\begin{align}
& P_D\left(w_{\tau}|s \in C\right) \\
=& P_D\left(w_{\tau} | \tau \in s, s \in C\right) P_D\left(\tau \in s|s\in C\right) 
\end{align}
Note, however, that we can estimate $ P_D\left(\tau \in s| s \in C \right) = \frac{|\tau \in C|}{|s \in C|}$ from data, such that 
\begin{align}
&P_D\left(w_{\tau}=\mu | \tau \in s, s \in C \right) P_D\left(\tau \in s| s \in C \right) \\
= &P_D\left(w_{\tau}=\mu | \tau \in s, s \in C \right) \frac{|\tau \in C|}{|s \in C|} \\
=&\sum_{s \in C} P_D\left(w_{\tau}=\mu |s, \tau \in s, s \in C \right) \frac{|\tau \in C|}{|s \in C|} \frac{1}{|\tau \in C|}  \\
=& \sum_{s \in C} \frac{g_{\mu}(s,\tau)}{|s \in C|}
\end{align}
We can drop the division by $|s \in C|$ whenever inference remains within a given $C$ such that $|s \in C|$ is constant. Otherwise, the divisor scales the aggregate tie strengths according to the size of the context $C$. That is, 
$$
g_{\mu,\tau}(C) \propto \sum_{s \in C} g_{\mu}(s,\tau)
$$
in contrast to our earlier measure of 
$$g_{\mu}(C,\tau)\coloneqq \sum_{\tau \in C} \frac{g_{\mu}(s,\tau)}{|\tau \in C|}$$

Edges in the (aggregate) conditioned substitution network $G^C$ represent the probability of substitution dyads within the context $C$. By contrast, edges in the original multigraph $G$ represent the contribution of $\mu$ to the semantic composition of $\tau$ independent of its frequency of occurrence. We further emphasize that, despite its simpler appearance, aggregate substitution necessitates two estimation steps, whereas compositional substitution requires only one.
In what follows, we refer to aggregate substitution.

\subsubsection{Structural disambiguation of semantics and syntax}\label{app:ch1:structural_identification}

Substitution relations that we extract from our language model convey information regarding syntax and semantics. 
Note first that a semantic relation is not a sufficient condition for a word to be able to replace another. Instead, for any reasonable computational language model, syntactic fit is a necessary condition for substitution. Words that replace each other, need to be able to function as the same part-of-speech (\cite{tenney_bert_2019}). As has been demonstrated in (\cite{hewittStructuralProbeFinding2019}), semantic relations are \textit{also} a necessary condition for a substitution tie. In consequence, substitution ties define both the semantics of a word, as well as its ability to function as a given part-of-speech. To make clean statements about the semantics of language use in corpus $D$, we need to identify sets of words where semantic similarity is not only necessary, but sufficient to obtain a substitution relation.

Substitution ties can be used to disambiguate between different syntactic functions of a word and its substitutions.
Generally, substitutions occur between words that function as the same part-of-speech in a given sentence. However, words can have several syntactic functions. For example, in the sentence \textit{the collared cat jumps across the river}, assume that the word \textit{collared} could be replaced by \textit{pet}. Since \textit{pet} is used as adjective, the substitution between \textit{collared} and \textit{pet} appears to cross a syntactic boundary: a tie exists between an adjective, and a word that is commonly used as a noun. 
Nevertheless, there are few such relations that do not significantly change the meaning of the sentence. For that reason, the likelihood that substitution ties cross syntactic boundaries in the above sense is low. 
In contrast, substitutions between words that have the same syntactic function are far more likely. For example synonyms, antonyms, hyponyms, or hypernyms may constitute valid substitutions in a given sequence.\footnote{Generally, the language model is less constrained in constructing valid sequences across aspects of meaning than across parts-of-speech. This is so, because the model learns syntax quicker than semantics. Antonyms, for example, may serve as valid substitutions if the training data includes similar sequences, where the meaning of the focal word is reversed. The viability of semantic classes beyond near-synonyms depends  on the sequence length in training.}

The relative sparsity of substitution ties that carry syntactic variation compared to ties that carry semantic content has consequences for the graph $G^C$. 
Words that belong in the same syntactic category are more likely to be related by substitution.
Crucially, if we aggregate across a set of sequences $C$ that span a sufficiently broad array of semantic relationships, then this aggregation integrates out the dependence that arises from any particular sequences and induces dense ties between semantically related words. At the same time, ties crossing syntactic boundaries remain relatively sparse. 
Consequentially, we can find clusters $S \subseteq G^C$ that roughly correspond to functional or syntactic units. That is, we use the variation in context in our corpus to identify syntactic boundaries, such that words within a given cluster have only semantic relations.
When analyzing substitution ties for semantics, we first specify which functional or syntactic aspects and which corresponding clusters are relevant, subsequently focusing on within cluster ties. The corresponding subgraph $S \subseteq G^C$  can be referred to as \textit{semantic substitution network}.

\subsubsection{A structural definition of semantics}\label{app:ch1:semantic_structures}
In this section, we use the structure of substitution ties to derive the \textit{semantic identity} of a focal token $\tau$. 

Recall that substitution ties in $G^C$ and its subgraphs are directed. For a focal token $\tau$, incoming ties of the form $g_{\mu,\tau}(C)$ and outgoing ties of the form $g_{\tau,\mu}(C)$ are both defining aspects of meaning. 

Incoming ties enumerate and give weight to the sense with which $\tau$ was used in our corpus. Outgoing ties, on the other hand, specify what meaning $\tau$ contributes to other words. 
We find that neither measure alone is sufficient to derive semantics of a word.


Consider a dyad of two words, $\mu$ and $\tau$.
We differentiate three dyadic constellations. If $g_{\mu,\tau}(C)=g_{\tau,\mu}(C)$, then the pairwise contribution of meaning of either word to the other is equal in $C$. We call this constellation of pairwise substitution \textit{symmetric}.
Take as example the words \textit{manager} and "leader". When "leader" is used in a non-academic business context, chances are the sentence in question would also allow the use of \textit{manager}. And, in most non-specialized publications, the use of \textit{manager} often refers to the role of a "leader".
Note, however, that the presence of symmetric semantic ties does not necessitate that two words are strong synonyms. First, the relationship may differ across contexts. In politics, a representative leader (for example a monarch) may be purposefully disconnected from the task of managing the countries' affairs. Second, even in a given context where $g_{\mu,\tau}(C)=g_{\tau,\mu}(C)$, it may very well be that $\mu$ and $\tau$ differ in how they relate to other words. That is, while both words tend to be similarly adequate substitutions to each other, each carries a distinct semantic identity stemming from other uses in the observed language. To be strong synonyms, $\mu$ and $\tau$ would not only be substitutions, but also feature a near-identical set if semantic ties to all other words. Instead, such symmetric substitutions are generally near-synonyms that retain a distinct semantic identity. In particular, the agreement between neighboring terms for two pairwise replacing words in $G^C$ is an attractive measure for the \textit{degree} of near-synonymy.

\begin{figure}[h]
    \centering
\efbox{
\begin{minipage}{0.3\textwidth}
\begin{tikzpicture}[node distance={25mm}, thick, main/.style = {draw, circle},xscale=1,yscale=1] 
\node[main] (1) {$\tau$}; 
\node[main] (2) [right of=1] {$\mu$}; 
\draw [bend right=-25,<-,line width=0.6mm] (1) to node [sloped, yshift=0.2cm,above,font=\small] {$g_{\mu,\tau}(C)$} (2);
\draw [bend right=-25,<-,line width=0.6mm] (2) to node [sloped, yshift=-0.2cm,below,font=\small] {$g_{\tau,\mu}(C)$} (1);
\end{tikzpicture}
\end{minipage}
\hfill
\begin{minipage}{0.3\textwidth}
\begin{tikzpicture}[node distance={25mm}, thick, main/.style = {draw, circle},xscale=1,yscale=1] 
\node[main] (1) {$\tau$}; 
\node[main] (2) [right of=1] {$\mu$}; 
\draw [bend right=-25,<-,line width=0.6mm] (1) to node [sloped, yshift=0.2cm,above,font=\small] {$g_{\mu,\tau}(C)$} (2);
\draw [bend right=-25,<-] (2) to node [sloped, yshift=-0.2cm,below,font=\small] {$g_{\tau,\mu}(C)$} (1);
\end{tikzpicture}
\end{minipage}
\hfill
\begin{minipage}{0.3\textwidth}
\begin{tikzpicture}[node distance={25mm}, thick, main/.style = {draw, circle},xscale=1,yscale=1] 
\node[main] (1) {$\tau$}; 
\node[main] (2) [right of=1] {$\mu$}; 
\draw [bend right=-25,<-] (1) to node [sloped, yshift=0.2cm,above,font=\small] {$g_{\mu,\tau}(C)$} (2);
\draw [bend right=-25,<-,line width=0.6mm] (2) to node [sloped, yshift=-0.2cm,below,font=\small] {$g_{\tau,\mu}(C)$} (1);
\end{tikzpicture}
\end{minipage}}
\captionsetup{justification=centering}
    \caption{\small Three dyadic structures.}
    \label{fig:c1:dyads}
\end{figure}
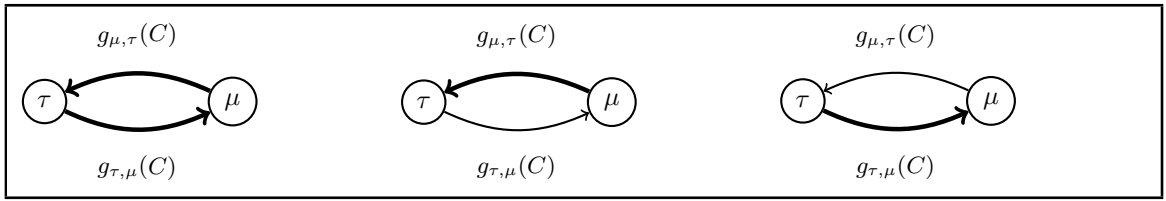

If, on the other hand, $g_{\mu,\tau}(C)>g_{\tau,\mu}(C)$, then $\tau$ was more often used in the sense of $\mu$, and $\mu$ is more determinant of $\tau$'s meaning in $C$. The situation is reversed if $g_{\mu,\tau}(C)<g_{\tau,\mu}(C)$. For example, in the sentence \textit{Jack is a leader with an inflated sense of his own importance for the company.}, the word $\tau=$"leader" could be replaced by the word $\mu=$\textit{narcissist}. The language model can deduce this contextual substitution since the semantic position of "leader" in this sentence is similar to descriptions of \textit{narcissists} in other parts of the corpus. Conversely, substitutions ties from $\mu=$\textit{narcissist} to $\tau=$"leader" are scarce, since there are few reasons to use the word narcissist in the sense of leader. 
Even if such ties exist for some sequences, they vanish compared to other semantic ties that determine the compositional substitution of "leader". Consequently, in the context $C$, we would conclude that \textit{narcissist} contributes more to the semantic identity of "leader" than the other way around. In other words, there is more narcissism in leadership than there is leadership in narcissism. $g_{\mu,\tau}(C)>g_{\tau,\mu}(C)$. Figure \ref{fig:c1:dyads} illustrates all three dyadic cases that will form the basis for our analysis of substantive elements of meaning.
To sum up, the semantic identity of a focal token is defined both by its outgoing and its incoming substitution ties across all neighboring tokens. While a symmetric pairwise relation implies near synonymy, we can only define strong synonyms through a comparison of all adjacent ties. Furthermore, non-symmetric substitution ties define when a word contributes meaning to other tokens, that is, when other tokens are used in the sense of the focal word. Whereas incoming ties give a sense of how a focal word was used in a given text, outgoing ties give rise to the meaning the word has and contributes within the corpus of consideration. 

\begin{figure}[h]
    \centering
\efbox{
\begin{minipage}{0.45\textwidth}
\begin{tikzpicture}[node distance={25mm}, thick, main/.style = {draw, circle},xscale=1,yscale=1] 
\path[step=1.0,black,thin,xshift=0.5cm,yshift=0.5cm] (2,0) grid (-3.5,0);
\node[main] (1) {$\tau$}; 
\node[main] (2) [below right of=1] {$\mu$}; 
\node[main] (3) [below left of=1] {$\rho$}; 
\draw [bend right=-25,<-,line width=0.4mm] (1) to node [sloped,above,font=\small] {$g_{\mu,\tau}(C)$} (2);
\draw [bend right=-25,<-,line width=0.4mm] (2) to node [sloped,above,font=\small] {$g_{\tau,\mu}(C)$} (1);
\draw [bend right=-25,<-,line width=0.4mm] (1) to node [sloped,above,font=\small] {$g_{\mu,\tau}(C)$} (3);
\draw [bend right=-25,<-,line width=0.4mm] (3) to node [sloped,above,font=\small] {$g_{\tau,\mu}(C)$} (1);
\end{tikzpicture}
\end{minipage}}
\captionsetup{justification=centering}
    \caption{\small An open triad.}
    \label{fig:c1:triads}
\end{figure}
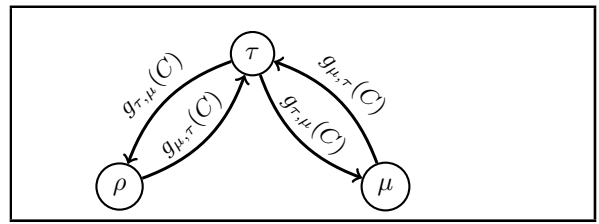

Next, consider a triad of three words, $\tau$, $\mu$, and $\rho$ in a semantic substitution network.
Generally, if $\tau$ has pairwise substitution relations to both $\mu$ and $\rho$, but $\mu$ and $\rho$ have no semantic connection, then we speak of an \textit{open triad}. In this triad, $\mu$ and $\rho$ are semantically distinct. $\tau$, however, is a near-synonym to either term. An \textit{open triad} indicates polysemy of $\tau$, or, more generally, it's use in distinct constellations of meaning. For example, the word \textit{chair} appears both as furniture and as role in an academic context.
Figure \ref{fig:c1:triads} shows an open triad. 
Note further that ties between elements of an open tried may be directed, in the sense that substitution relations are skewed more strongly towards any of the three elements. For example, if both $\mu$ and $\rho$ are more likely substitutions for $\tau$, but not for each other, then they constitute separate elements of $\tau$'s semantics whenever it is used in the corpus. Conversely, if $\tau$ strongly replaces either $\mu$ or $\tau$, then these words are used in the sense of $\tau$. In other words, $\tau$ constitutes a common semantic aspect between these words. Consequently, such triads are useful to define the meaning of $\tau$ itself, namely, all the aspects of meanings it confers on other, possibly distinct tokens.

In conclusion, both immediate substitution ties and the collection of adjacent words are crucial to identify meaning from a semantic network. In addition, non-symmetric ties can be employed to derive a deeper understanding of semantics, with outgoing ties capturing the meaning that a focal token confers upon another word and ultimately a sentence in a given text.

\subsection{Derivation of context measures}
In the prior sections, we have seen that the semantic network can easily be conditioned on a set of sequences. 
There is no reason, however, to restrict contextual inference to exogenous sets of sentences. Instead, we may wish to infer which contexts are relevant, for example for a given focal term.
Furthermore, we may want to condition the semantic substitution network on a context defined by logical conditions, for example all sentences about the topic "leadership".

In each case, we require some measure of contextual relevance. The objective of this section is to illustrate how such measure can be defined from the language model.

Since all sequences in $D$, and thus information on all positions $s_i$ are available, we can also derive context measures directly from the corpus. This is the co-occurrence approach. For example, consider any sequence $s \in D$. In a sequence of length $|s|$, for a word $\rho$ that occurs $k$ times, let $$q_{\rho}^*(s)=|\rho \in s|=k$$
Hence, $\frac{{q}_{\rho}^*(s)}{|s|}$ simply denotes how much of the sequence $s$ is made up of $\rho$. If, for example, $\rho$ occurs once in $s$, we naturally have $\frac{q_{\rho}^*(s)}{|s|}=\frac{1}{|s|}$, corresponding to the probability that any randomly selected position in $s$ is the word $\rho$.\footnote{These co-occurrence distributions $q_s^*$ 
are available from $G$. Recall that a word $\rho$ that occurs $k$ times in $s$ implies that $\sum_{\gamma} g_{\gamma,\rho}(s) = k$ where $\gamma \in \Omega$, since incoming ties of each occurrence sum to unity. Similarly, the length $|s|$ of sequence $s$ can be found by $\sum_{\delta}\sum_{\gamma} g_{\gamma,\delta}(s)=|s|$. That is
$$
\frac{q_{\rho}^*(s)}{|s|} = \frac{\sum_{\gamma} g_{\gamma,\rho}(s)}{\sum_{\delta}\sum_{\gamma} g_{\gamma,\delta}(s)}  = \frac{\sum_{\gamma} g_{\gamma,\rho}(s)}{|s|} = \frac{k}{|s|}
$$}
However, we have already seen how our language model improves upon frequency based approaches such as a context measure based on co-occurrences. We will therefore focus on deriving contextual measures from the language model itself.

\subsubsection{Substitution measures of context}

Prior discussion has noted how substitution ties, outgoing ties in $G$, are based on the higher-order relationships that the language model has learned from the corpus $D$. We have asserted that this information is superior to relationships derived from co-occurrence methods, since co-occurrences draw connections  only insofar that tokens actually appear in writing, not, however, when they \textit{could have been written.} 
$q_s^*$ is a co-occurrence measure of context. It is the likelihood of randomly picking an actually occurring word from a sequence $s$. A more stable measure, and one that incorporates more information from the corpus at hand, would take words into account that \textit{could} appear if the internal language of the text indicates it. For example, if a sentence includes the verb \textit{speaks}, then this sentence should likely also be considered relevant if we are interested in the contextual term \textit{talks} (at least if \textit{talks} were are near-synonym in the focal sequence). This determination of relevance is precisely what substitution distributions from \textit{BERT} indicate.

\subsubsection{Joint substitution distributions} \label{app:ch1:dyadic_context}

The goal of this section is to derive measures that represents the occurrence of two substitution ties $w_{\tau}=\mu$ and $w_{\delta}=\rho$, with $\tau$ and $\delta$ occurring, and $\mu$ and $\rho$ substituting. Once derived, these measured can be used to get to different kinds of aggregations.

Whereas dyadic substitution ties between two tokens are defined for specific occurrences, and result in a single probability distribution, we instead face a number of contextual terms and, in turn, many different substitution distributions.
Thus, it is no longer trivial to deal with positional information, nor is the matter of aggregation as straightforward. 
Using assumption \eqref{app:ch1:ass2}
$$
P\left(w_i, w_j |s\right)=P\left(w_i, w_j |w_{-i,j}, s_i, s_j\right) = P_D\left(w_i|s\right)P_D\left(w_j|s\right)
$$

Assume now that each word occurs at most once in $s$. Then, the joint probability above can be used as a measure for context. Recall that we denoted the event $w_{\tau}= \cup_i \{w_i, s_i=\tau\}$. 
First, the joint probability that $\tau$ is substituted by $\mu$ and $\delta$ is substituted by $\rho$ is given by
$$
P\left(w_{\tau}=\mu, w_{\delta}=\rho |s\right)=\sum_{(i,j)} P_D\left(w_i=\mu, s_i=\tau|s\right)P_D\left(w_j=\rho, s_j=\delta|s\right)
$$
where the sum is nonzero for only a single element. 
For two fixed dyads and a given sequence $s$, this measure is straightforward to calculate. We will make use of this formulation to calculate substitution relationships between $\rho$ and $\tau$ under the condition that the sentence $s$ includes a substitution relationship between $\mu$ and $\tau$ in section \ref{app:ch1:context_sub} below. We will calculate
\begin{align}
    q_{\rho, \delta}^{\mu, \tau}(s) \defeq P_D\left(w_i=\mu|s_i=\tau, s_{-i}\right)P_D\left(w_j=\rho|s_j=\delta, s_{-j}\right)
\end{align}
which is zero if either $\tau$ or $\delta$ do not occur in $s$.

Suppose, however, that we are interested in the context of an instance where $\rho$ is replacing any other term, say, 
 $w_{.\setminus \tau}= \cup_i \{w_i, s_i \in \Omega_{\setminus \tau} \}$. 
 Then, taking into account the first dyad of $\mu$ substituting for $\tau$, we seek 
$P\left(w_{\tau}=\mu, w_{.\setminus \tau}=\rho |s\right)$. In other words, where the above measure specifies four tokens $\mu$,$\tau$,$\rho$, and $\delta$, we seek a measure where at least one token is a free parameter. Even if we maintain the assumption that each occurrence is unique, $\rho$ might substitute for several occurrences in $s$, such that we need to consider a set of position pairs $(i,j)$ instead of a single, well-defined pair. 

That is, we factorize
$$
P\left(w_{\tau}=\mu, w_{.\setminus \tau}=\rho |s\right) = P\left(w_{.\setminus \tau}=\rho | w_{\tau}=\mu, s\right) P\left(w_{\tau}=\mu |s\right)$$
given a unique occurrence of $\tau$, or equivalently, our scheme of marginalizing over distinct occurrences. Applying conditional independence, we have
$$
P\left(w_{\tau}=\mu, w_{.\setminus \tau}=\rho |s\right) = P\left(w_{j \neq i}=\rho | s\right) P_D\left(w_{i}=\mu |s, s_i=\tau \right)$$

where $\{w_{j \neq i}=\rho \}$ is the event of $\rho$ substituting in $s$ at any other position than $i$, that is
$\{w_{j \neq i}=\rho\} \eqdef \cup_{j \neq i} \{w_j=\rho, s_j=\delta\}=\{\cup_{\delta \in \Omega \setminus \rho} w_{\delta}=\rho\}$. Here, note that 
$\cup_{j \neq i} \{w_j, s_j\}$ is a random element on $\Omega*\Omega$.
The corresponding probability distribution is $P\left(\cup_{j \neq i} \{w_j, s_j\} | s\right)$. For a given $s$, we have $P\left(\cup_{j \neq i} \{w_j, s_j\} | s\right)=P\left(\cup_{j \neq i} w_j |  s\right)$. The probability distribution is
\begin{align*}
P\left(\cup_{j \neq i} w_j | s\right) &= \sum_{j \neq i} P\left(w_j|s\right) - \sum_{j,k \neq i} P\left(w_j, w_k |s\right) + \sum_{j,k,v} \ldots \\
&= \sum_{j \neq i} P\left(w_j|s\right) - \sum_{j,k \neq i} P\left(w_j|s\right)P\left(w_k|s\right) + \sum_{j,k,v} \ldots
\end{align*} by the inclusion-exclusion formula. Then
\begin{align*}
P\left(w_{\tau}=\mu, w_{.\setminus \tau}=\rho |s\right) = \left[ \sum_{j \neq i} P_D\left(w_j|s\right) - \sum_{j,k \neq i} P_D\left(w_j|s\right)P_D\left(w_k|s\right) + \sum_{j,k,v} \ldots\right] P_D\left(w_{i}=\mu |s, s_i=\tau \right) \\
=\sum_{j \neq i} P_D\left(w_j|s\right) P_D\left(w_{i}=\mu |s, s_i=\tau \right) - \left[ \sum_{j,k \neq i} P_D\left(w_j|s\right)P_D\left(w_k|s\right) - \sum_{j,k,v} \ldots\right] P_D\left(w_{i}=\mu |s, s_i=\tau \right)
\end{align*}
Given that probabilities are at most one, the leftwise multiplications in the above formula are a magnitude larger than subsequent elements. If the model is trained to an acceptable level of entropy, we have an upper bound with relatively small error:
\begin{align*}
P\left(w_{\tau}=\mu, w_{.\setminus \tau}=\rho |s\right) 
\leq \sum_{j \neq i} P_D\left(w_j|s\right) P_D\left(w_{i}=\mu |s, s_i=\tau \right)
\end{align*}
which we will use as approximate measure of context:
\begin{align*}
P\left(w_{\tau}=\mu, w_{.\setminus \tau}=\rho |s\right) 
&\approx \sum_{j: s_j \neq \tau} P_D\left(w_{j}|s\right) P_D\left(w_{t}=\mu |s \right) \\ &= P_D\left(w_{. \setminus \tau}|s\right) P_D\left(w_{t}=\mu |s \right) \\
&\eqdef q_{\rho}^{\mu, \tau}(s)
\end{align*}
where, for notational simplicity, we define 
\begin{align} \label{eqn:app:qdefinition}
    q_{\rho}(s) &= P_D\left(w_{.}=\rho |s\right) =\sum_{\delta} g_{\rho,\delta}(s) \\
    q_{\rho}(s\setminus \tau) &= P_D\left(w_{.\setminus \tau}=\rho \right) =\sum_{\delta \neq \tau} g_{\rho,\delta}(s)
\end{align}
leading to the measure
\begin{align}\label{eqn:app:jointApprox}
q_{\rho}^{\mu, \tau}(s)= q_{\rho}(s\setminus \tau)g_{\mu, \tau}(s)
\end{align}

For a given context $C$ we will use the same occurrence-based aggregation as in the substitution measure, that is
\begin{align*}\label{eqn:app:jointApproxAgg}
P_D\left(w_{\tau}=\mu, w_{.\setminus \tau}=\rho |C\right) 
&\approx \sum_{s_j=\tau \in C}\sum_{j: s_j \neq \tau} P_D\left(w_{j}|s_i=\tau, s \in C \right) P_D\left(w_{i}=\mu | \tau \in s \in C \right)P\left(\tau \in s|s\in C\right)
\\ &= \sum_s P_D\left(w_{. \setminus \tau}|s\right) P_D\left(w_{t}=\mu |s \right) \frac{|\tau \in s|}{|s \in C|} \\
&= \sum_s g_{\mu, \tau}(s) q_{\rho}(s\setminus \tau) \frac{1}{|s \in C|} \\
&\eqdef q_{\rho}^{\mu, \tau}(C)
\end{align*}
As before, dropping the assumption that each word occurs once in a sentence, we instead marginalize over occurrences instead of sequences, without changing the rest of the equation.

Due to size of the vocabulary, the magnitudes of probabilities are generally small, such that the approximation error in equation \ref{eqn:app:jointApprox} is minor. However, using usual occurrences measures as inspiration,  section \ref{app:ch1:randomContextElement} develops an analogous measure from the language model, which turns out to be a lower bound to the above probability.

For now, we apply our approximate measures to finding contextual substitutions, context to context relationships, and a scheme to condition on contextual words instead of specific sequences.

\subsection{Context substitution network}\label{app:ch1:context_sub}

The measures developed thus far give the likelihood of a contextual word, or a contextual relationship together with the substitution dyad $\mu, \tau,$ or simply the substitution of $\mu$. The latter acts as a weight on the random element $w_{.}=\rho$, which we defined as context.

Fixing $\mu$ we can derive a network $QS$, where the tie between $\rho$ and $\delta$ is given by $q^{\mu}_{\rho, \delta}$ from  \ref{eqn:app:condContext}, or fixing the dyad $\mu, \tau$, ties in $QS$ between $\rho$ and $\delta$ are given by $q^{\mu,\tau}_{\rho, \delta}$ from \ref{eqn:app:doubledyadicContext}. Similar to the substitution network, this context substitution network is a multi-graph, unless conditioned on a specific set $C$.

The interpretation is similar to the substitution network $G$: we get the probability of $\rho$ substituting for $\delta$ in $C$, now however with the additional condition that this substitution happens anywhere in the context of $\mu$ substituting $\tau$.

\subsection{Context element network}

Thus far, the context measures gave relations between occurring token (e.g. $\tau$) and substituting token (e.g. $\mu$), possibly anywhere in the context of a relationship of interest. By factorizing further, we can expand our previous measure to two substituting contextual tokens. 
Specifically, imagine the joint likelyhood that $\rho$ replaces some word in $s$, and $\gamma$ replaces \textit{another word in s}. We are looking for a probability akin to 
$$
P\left(w^1_{.}=\rho,w^2_{.} = \gamma, w_{\tau} = \mu|s\right)
$$
where $w^1_{.}$ and $w^2_{.}$ indicate the substitution distributions of a pair of words in $s$. Under the independence assumptions, we can again find an approximation 
\begin{align*}
&P\left(w^1_{.}=\rho,w^2_{.} = \gamma, w_{\tau} = \mu|s \right) \\
=& \sum_{\substack{w_{\delta}: s_\delta \neq \psi, \tau \\ w_{\psi}: s_\psi \neq \delta, \tau}} P_D\left(w_{\psi}=\gamma|w_{\delta}=\rho, w_{\tau} = \mu, s \right)  P_D\left(w_{\delta}=\rho|w_{\tau} = \mu, s\right) P_D\left(w_{\tau} = \mu |s \right) \\ \approx&  P_D\left(w_{\tau}=\mu |s \right)\sum_{\delta: \neq \tau} P_D\left(w_{\delta}| s\right) \sum_{\psi \neq \delta, \tau} P_D\left(w_{\psi}=\gamma| s \right) \\
\\ =&  g_{\mu, \tau}(s) \sum_{\delta: \neq \tau} g_{\rho, \delta}(s) \sum_{\psi \neq \delta, \tau} g_{\gamma, \psi}(s) \\
\defeq & q_{\rho, \gamma}(s\setminus \tau) g_{\mu, \tau}(s)
\end{align*}

where the approximation arises from the assumption that words are unique in the sentence, and positions are conditionally independent. In the fourth line, it becomes apparent that the computational implementation needs to expand from a given dyad $\mu, \tau$ to occurrences yielding a dyad involving $\rho$, to then distinct occurrences giving a dyad with $\gamma$. Here, an implementation taking advantages of a graph database allows performant queries with successive matches.

In turn, we define the context element network $Q$, where ties are defined between contextual tokens $\rho$ and $\gamma$, with the condition of $\mu$ substituting for $\tau$.
Denote these context element ties by $e^{\mu,\tau}_{\rho, \gamma}$ to distinguish them from the context substitution network.

\subsection{Bidirectional context: substitution and occurrence}\label{app:ch1:context_bidirectional}

We have spent considerable time developing a substitution measure of context that relies on the language model, that is
$$q_{\rho}(s) = P\left(w_{.}=\rho |s\right) =\sum_{\delta} g_{\rho,\delta}(s)$$

which, in terms of the multigraph G, differs from the occurrence measure merely by the tie direction, that is
$$q^*_{\rho}(s) =  \sum_{\delta} g_{\delta, \rho}(s) = |\rho \in s|$$

We  can mix these two notions, setting the probability equal to $1$ if $\rho$ occurs, or setting it to the likelihood of such occurrence (as determined by our language model) otherwise. This mixture maximizes recall: we find context sequences where words appear, or could appear with high likelihood. Given the binary nature of $q^*_{\rho}(s \setminus \tau)$, a simple combination of the two measures suffices, that is,
$$q_{\rho}(s)^{\text{bidirectional}}=\min\left(q^*_{\rho}(s)+ q_{\rho}(s), 1\right)$$

All other contextual measures can be transformed accordingly.
\subsection{Conditioning $G$ on a set of contextual words}\label{app:ch1:context_element}
In the main part, we clarified how $G$ can be conditioned on an arbitrary set of sequences $C$. However, such a set needs to be defined. 
In particular, suppose we wish to examine the substitution ties of token $\mu$ for a context set specified as follows
$$
C \defeq \{\text{all sentences where $\rho$ occurs in context.}\}
$$
In our empirical application, for example, we derive the substitution ties of the word $\mu=$"founder" for contexts that include specific firms like $\rho=$\textit{facebook} or $\rho=$\textit{microsoft}.
The desired context $C$ could also include several words, for example, $\rho_1=$\textit{microsoft}, $\rho_2=$\textit{gates}, $\rho_3=$\textit{ballmer}. 

Let $\Lambda = \{\rho_1, \rho_2, \ldots\}$ be the contextual words of interest.
Recall that the aggregate substitution distribution of $\mu$ across all $\tau$ in $C$ is given by
$$
g_{\mu, \tau}(C) = P_D\left(w_{\tau}=\mu|s \in C\right)
$$
we will instead derive a measure 
$$
P\left(w_{\tau}=\mu| ~\text{$s$ includes at least one $\rho_i \in \Lambda$}\right)
$$
and denote it by $g_{\mu, \tau}(\Lambda)$. 
Similar to the prior sections, we again need to derive a likelihood for each sequence $s$ in order to sum up occurrences. Instead of conditioning on the event that $s \in C$ for some $C$, which required an assessment of $P(s|s\in C)$ to marginalize over, we now consider the more complex probability
$$
P\left(\text{$s$ includes at least one $\rho_i \in \Lambda$}\right)$$

We can now employ our aforementioned measures of context. Note first the deterministic co-occurrence measure can give an exact result: $\underline{q}_s^*(\rho_i)$ is positive, iff $\rho_i$ appears in $s$. Since we will consider substitution ties between some token $\mu$ and $\tau$, a correct specification of context requires the use of $q^*_{\rho_i}(s \setminus \tau)$.
In turn, $P\left(\text{$s$ includes at least one $\rho_i \in \Lambda$}\right)$ is one whenever at least one term appears in $s$. That is
$$
P\left(\text{$s$ includes at least one $\rho_i \in \Lambda$}\right) = 1 \Leftrightarrow \exists \rho_i \in \Lambda:  q^*_{\rho_i}(s \setminus \tau)>0$$

Note also that since $q^*_{\rho_i}(s \setminus \tau) \geq 1$ if it is nonzero, we can equivalently consider the condition
$$\sum_{\rho_i} q^*_{\rho_i}(s \setminus \tau) = \sum_{\rho_i}P\left(s_i=\rho_i ~\text{for a uniformly random}~ s_i \in s \right) \geq 1$$

The prior formulation allows us again to transfer the idea to the substitution measure of context, enriching the set of sequences we may consider with positive weight. Then, the likelihood can be specified by the probability that any $\rho_i \in \Lambda$. In other words, we use the previously specified measure 
$$q_{\rho_i}(s)=\frac{\sum_{\gamma} g_{\rho_i,\gamma}(s)}{\sum_{\delta}\sum_{\gamma} g_{\gamma,\delta}(s)} =  \frac{\sum_{\gamma} g_{\rho_i,\gamma}(s)}{|s|}$$
where $|s|$ is the length of the sequence. 
Recall that $q_{\rho_i}(s \setminus \tau)$ gives the likelihood that $\rho_i$ is appropriate in $s$ on a position other than one where $\tau$ occurs. Assuming that these occurrences are independent, we can derive a substitution based weight for a sequence $s$ as
$$P\left(\text{$s$ includes at least one $\rho_i \in \Lambda$}\right)=\sum_{\rho_i}P\left(s_i=\rho_i ~\text{for a uniformly random}~ s_i \in s \right)$$ 
where now\footnote{Note that
$
\sum_{\rho_i} q_{\rho_i}(s \setminus \tau)=\frac{\sum_{\rho_i}\sum_{\gamma \neq \tau} g_{\rho_i,\gamma}(s)}{ \sum_{\delta}\sum_{\gamma \neq \tau} g_{\gamma,\delta}(s)} \leq 1
$ and if $\Lambda = \Omega$ then $\sum_{\rho_i} q_{\rho_i}(s \setminus \tau)=\frac{\sum_{\rho \in \Omega}\sum_{\gamma \neq \tau} g_{\rho_i,\gamma}(s)}{ \sum_{\delta \in \Omega}\sum_{\gamma \neq \tau} g_{\gamma,\delta}(s)}=1
$} $$\sum_{\rho_i}P\left(s_i=\rho_i ~\text{for a uniformly random}~ s_i \in s \right) = \sum_{\rho_i} q_{\rho_i}(s \setminus \tau)$$
as likelihood weight of $s$. 
The advantage is apparent: $P\left(\text{$s$ includes at least one $\rho_i \in \Lambda$}\right)$ can be larger than 0 even if no $\rho_i \in \Lambda$ actually occurs. In turn, the measure identifies contexts via substitution ties and is therefore based on the language model's assessment of contextual relevance.

As bidirectional measure, we use
$$P\left(\text{$s$ includes at least one $\rho_i \in \Lambda$}\right)=\min\left(\sum_{\rho_i} q^*_{\rho_i}(s \setminus \tau)+\sum_{\rho_i} q_{\rho_i}(s \setminus \tau), 1\right)$$

Finally, we use any version of these weights to derive the re-weighted substitution ties
$$
g_{\mu,\tau}(\Lambda) \eqdef \sum_{s \in D}g_{\mu,\tau}(s)P\left(\text{$s$ includes at least one $\rho_i \in \Lambda$}\right)
$$
Note that if we use the occurrence based notion, we have
$g_{\mu,\tau}(\Lambda)=g_{\mu,\tau}(C)$ where $C$ includes all sentences with the desired words. Using substitutions, we derive a richer notion of context, one that includes sequences where tokens $\rho_i$ have a high likelihood of occurrence, because the language model judges them to be a good fit in terms of syntax and semantics.

\subsubsection{Random context element as lower bound joint distribution}\label{app:ch1:randomContextElement}
We have previously used an approximation, arguing that the approximation error is small. We now develop a related but exact formulation. Consider an expected distribution of context, where we assume that one $j$ is picked at random, that is, from a uniform distribution $j : U(1,|s|)$ where $|s|$ is the number of elements in $s$, and 
$$\underline{w}_{\delta} \defeq E_j \left(w_j\right)$$
which, for some given occurring token $\delta$ is simply
$$P\left(\underline{w}_{\delta}|s\right)= \sum_{j} \frac{  P\left(w_j|s\right)}{|s|}$$
Alternatively, we can focus on the substituting token $\rho$, allowing substitution for any occurring element
$$P\left(\underline{w}_{.}=\rho|s \right) \defeq \sum_{\delta \neq \rho} \frac{  P\left(w_{\delta}=\rho|s\right)}{|s|}$$
We further find that the substitution network $G$ can be used to derive these measures efficiently, giving the following measures for the token $\rho$ or for a  specific dyad $\delta, \rho$, and another occuring token $\tau$
\begin{align}
    \underline{q}_{\rho, \delta}(s)&=P\left(\underline{w}_{\delta}=\rho|s \right) =\frac{ g_{\rho,\delta}(s)}{\sum_{\phi}\sum_{\gamma} g_{\gamma,\phi}(s)} \\
    \underline{q}_{\rho, \delta}(s  \setminus \tau) &=P\left(\underline{w}_{\delta \setminus \tau}=\rho|s \right) =\frac{ g_{\rho,\delta}(s)}{\sum_{\phi}\sum_{\gamma \neq \tau} g_{\gamma,\phi}(s)} \\
    \underline{q}_{\rho}(s) &= P\left(\underline{w}_{.}=\rho |s\right) =\frac{\sum_{\delta} g_{\rho,\delta}(s)}{\sum_{\phi}\sum_{\gamma} g_{\gamma,\phi}(s)} \\
    \underline{q}_{\rho}(s\setminus \tau) &= P\left(\underline{w}_{.\setminus \tau}=\rho \right) =\frac{\sum_{\delta \neq \tau} g_{\rho,\delta}(s)}{\sum_{\phi}\sum_{\gamma \neq \tau} g_{\gamma,\phi}(s)}
\end{align}

with $\underline{w}_{\delta}$ defining the random element for a focal occurring token $\delta$ and $\underline{w}_{.}$ for any occurring token. In either case, we define the event ignoring a position in $s$ where the token $\tau \neq \rho$ occurs. This merely changes the denominator.

Note the similarity to a conventional co-occurrence measure. As above, if $\tau$ occurs $k$ times in $s$, then co-occurrence can be calculated from the reverse ties in $G$
$$
\underline{q}_{\rho}^*(s) = \frac{\sum_{\gamma} g_{\gamma,\rho}(s)}{\sum_{\delta}\sum_{\gamma} g_{\gamma,\delta}(s)}  = \frac{\sum_{\gamma} g_{\gamma,\rho}(s)}{|s|} = \frac{k}{|s|}
$$

\subsubsection{Contextual events and substitution dyads}

Having defined a suitable aggregation, we can now return to the joint distribution of two dyads. 

Assume again that $\tau$ occurs once in $s$, e.g., $s_i=\tau$. Otherwise, if $s_i=s_z=\tau$, consider each occurrence as a separate sequence.
We now have for a given $\mu$ and $\tau \in s$, as well as a separate dyad $\rho$ and $\delta \in s$
$$P\left(w_{\tau}=\mu,\underline{w}_{\delta \setminus \tau}=\rho  \right|s)= P\left(w_{\tau}=\mu|s\right) P\left(\underline{w}_{\delta \setminus \tau}=\rho  | s\right)
$$
giving us the joint probability of a certain substitution tie, and the event of another substitution tie in its context.

In particular, for a given $C$ by assumption \eqref{app:ch1:ass1}
$$
P\left(\underline{w}_{\delta \setminus \tau}|s \right) = P\left(\underline{w}_{\delta \setminus \tau}|s, s \in C \right)
$$
And
\begin{align*}
    P\left(w_{\tau}=\mu,\underline{w}_{\delta \setminus \tau}=\rho | s \in C\right) 
    =& \sum_{s \in C} P\left(w_{\tau}=\mu,\underline{w}_{\delta \setminus \tau}=\rho |s, s \in C\right)  P\left(s|s\in C\right) \\
    =&
    \frac{1}{|s \in C|} \sum_{s \in C} P\left(w_{\tau}=\mu,\underline{w}_{\delta \setminus \tau}=\rho |s, s \in C\right) \\
    =&  \frac{1}{|s \in C|} \sum_{s \in C} g_{\mu,\tau} (s) \frac{g_{\rho,\delta}(s)}{\sum_{\phi}\sum_{\gamma \neq \tau} g_{\gamma,\phi}(s)}
\end{align*}
similarly for any occurrence relevant to $\rho$
\begin{align*}
    P\left(w_{\tau}=\mu,\underline{w}_{. \setminus \tau}=\rho | s \in C\right) 
    =  \frac{1}{|s \in C|} \sum_{s \in C} g_{\mu,\tau} (s) \frac{\sum_{\delta \neq \tau} g_{\rho,\delta}(s)}{\sum_{\phi}\sum_{\gamma \neq \tau} g_{\gamma,\phi}(s)}
\end{align*}
Where the denominator $|s_{\setminus \tau}| = \sum_{\phi}\sum_{\gamma \neq \tau} g_{\gamma,\phi}(s)$ gives the length of the sequence without the focal occurrence(s) of $\tau$.
Define 
\begin{align}
    \label{eqn:app:dyadicContext}\underline{q}^{\mu,\tau}_{\rho}(C) = \frac{1}{|s \in C|} \sum_{s} g_{\mu,\tau}(s) \underline{q}_{\rho}(s\setminus \tau)   =\frac{1}{|s \in C|} \sum_{s} g_{\mu,\tau}(s) \frac{\sum_{\gamma \setminus \tau} g_{\rho, \gamma}(s)}{|s_{\setminus \tau}|} 
\end{align}
as the \textit{dyadic context} for a given dyad $\mu, \tau$, a distribution of contextual words across $\Omega$.
Note that if we analyze within a context $C$, we can surpress the normalization constant $\frac{1}{|s \in C|} $

Finally, for the same dyad $\mu, \tau$, the probability that \textit{a random contextual word is $\delta$, and is being substituted by $\rho$}, is given by 
\begin{align}
    \label{eqn:app:doubledyadicContext}\underline{q}^{\mu,\tau}_{\rho, \delta}(C) = \frac{1}{|s \in C|} \sum_{s} g_{\mu,\tau}(s) \underline{q}_{\rho, \delta}(s\setminus \tau)   = \frac{1}{|s \in C|} \sum_{s} g_{\mu,\tau}(s) \frac{g_{\rho, \delta}(s)}{|s_{\setminus \tau}|}
\end{align}

Analogously, define again the occurrence version of these measures as
$$
\underline{q}^{\mu,\tau *}_{\rho}= \frac{1}{|s \in C|} \sum_{s} g_{\mu,\tau}(s) \underline{q}_{\rho}^*(s\setminus \tau)   =\frac{1}{|s \in C|} \sum_{s} g_{\mu,\tau}(s) \frac{\sum_{\gamma} g_{\gamma,\rho}(s)}{|s_{\setminus \tau}|}
$$

\subsubsection{Conditional formulation}

Note also that given assumption \eqref{app:ch1:ass2},
$$
P\left(\underline{w}_{\delta \setminus \tau}| w_{\tau}=\mu, s \right) = P\left(\underline{w}_{\delta \setminus \tau}|s \right) 
$$
This allows us to use $P\left(w_{\tau}=\mu|s\right)$ as a relevance weight\footnote{To see this in more detail: $P\left(\underline{w}_{\delta \setminus \tau}| w_{\tau}=\mu, s \in C \right)  = \sum_{\tau \in s \in C} P\left(\underline{w}_{\delta \setminus \tau}| w_{\tau}=\mu, s, s \in C \right) P\left(s|  w_{\tau}=\mu, s\in C\right)
    =\sum_{\tau \in s \in C} P\left(\underline{w}_{\delta \setminus \tau}| w_{\tau}=\mu, s, s \in C \right) \frac{P\left(s,  w_{\tau}=\mu| s\in C\right)}{P\left(w_{\tau}=\mu| s\in C\right)} 
    =\sum_{\tau \in s \in C} 
    P\left(\underline{w}_{\delta \setminus \tau}| w_{\tau}=\mu, s, s \in C \right) \frac{P\left(w_{\tau}=\mu| s, s\in C\right)P\left(s| s\in C\right)}{P\left(w_{\tau}=\mu| s\in C\right)} 
    = \frac{1}{P\left(w_{\tau}=\mu| s\in C\right)} \sum_{\tau \in s \in C} P\left(\underline{w}_{\delta \setminus \tau} | s\right) P\left(w_{\tau}=\mu|s\right)P\left(s| s\in C\right)$.} for a collection of sequences $s$ in a context $C$.
$s \in C$ are disjoint, applying the law of total probability and Bayes' formula, we have 
\begin{align*}
    P\left(\underline{w}_{\delta \setminus \tau}| w_{\tau}=\mu, s \in C \right)  =& \frac{P\left(w_{\tau}=\mu,\underline{w}_{\delta \setminus \tau}=\rho | s \in C\right) }{P\left(w_{\tau}=\mu| s\in C\right)}  \\
    =& \frac{1}{|s \in C|}\frac{1}{P\left(w_{\tau}=\mu| s\in C\right)} \sum_{\tau \in s \in C} P\left(\underline{w}_{\delta \setminus \tau} | s\right) P\left(w_{\tau}=\mu|s\right)
\end{align*}
Recall that 
$P\left(w_{\tau}=\mu| s\in C\right) = \sum_{s \in C} g_{\mu,\tau} (s) \frac{1}{|s \in C|}$
and hence
\begin{align}
    \label{eqn:app:condContext}
P\left(\underline{w}_{\delta \setminus \tau}| w_{\tau}=\mu, s \in C \right)=   \sum_{s \in C} \frac{g_{\mu,\tau} (s)}{\sum g_{\mu,\tau} (k)} \frac{\sum_{\delta \neq \tau} g_{\rho,\delta}(s)}{\sum_{\phi}\sum_{\gamma \neq \tau} g_{\gamma,\phi}(s)}
\end{align}

The difference between equation \ref{eqn:app:condContext} and $\underline{q}^{\mu,\tau}_{\rho}(C)$ from equation \ref{eqn:app:dyadicContext} is that the former is conditional on the dyad $(\mu,\tau)$ occurring, and does not depend on whether $P\left(w_{\tau}=\mu| s\in C\right)$ is high or low.
The difference is similar the one between aggregate and compositional substitution, with the added complication of approximating a joint distribution.

\subsection{Further networks}

\subsubsection{Bidirectional and min / max substitution}
Not every analysis requires directional information. We may use three additional symmetric measures.
Bidirectional substitution, defined as $\hat{G}$ where  $\hat{g}^C_{\mu,\tau}=\frac{g^C_{\mu,\tau}+g^C_{\tau,\mu}}{2}$ specifies when words are mutual substitutions to each other.
Similarly, we define max substitution $\overline{G}$ as  $\overline{g}^C_{\mu,\tau}=\max \left(g^C_{\mu,\tau},g^C_{\tau,\mu}\right)$ and min substitution $\underline{G}$ as  $\underline{g}^C_{\mu,\tau}=\min \left(g^C_{\mu,\tau},g^C_{\tau,\mu}\right)$.
We proceed similarly for $R$.

\subsubsection{Entropy network}

Entropy of a probability distribution denotes the expected surprise or information that could be gained by observing a realization. To the degree that the distribution focuses its weight on a single term, for example, one would not learn additional information if that term is realized. If, however, the distribution is uniform, then the information gain by a hypothetical observation is maximized.
Hence, entropy describes the uncertainty of the probability distribution, a fact we can use to construct an entropy network.

Let $H_D(w_{\tau}| \tau \in s, s)$ be the entropy associated with predicting the substitution of $\tau$ based on the edges in $G$.
For each substitution $\mu$ that has a sufficiently high probability to be selected, we can therefore denote $h_{\mu,\tau}(s)$ as a tie giving the uncertainty of $\mu$ replacing $\tau$. With slight abuse of notation, we also call this graph $H$. Note that for a given sequence and focal token, each alternative word receives a tie of the same strength. This measure can be conditioned on $C$ in the same manner as above.

The network so constructed can be used to find clusters of terms that replace each other in situations of high uncertainty. In other words, whether the context implies a wide array of possible alternatives, or whether the context leads to a more specific connection.
Central words in the entropy network are universal or general, while non-central words are specific.

We briefly present two additional measures, that may be useful for specific research questions. These measures do not have a formal probabilistic interpretation, but are compound measures.

\textit{Certainty} is a compound measure defined as $\frac{g_{\mu,\tau}(s)}{h_{\mu,\tau}(s)}$. The measure similar to substitution, but weighs higher situations in which a substitution happens with fewer alternatives.

\textit{Unconventionality} is a similar compound measure. It adds to the above the idea that a word that stands out in situations of relatively high entropy, is a word that is unconventional compared to similar uses in the context. It is defined as $-\frac{h_{\mu,\tau}(s)}{\log(g_{\mu,\tau}(s))}$ and gives high weight to words that are certain relative to high entropy.
Use of these measures is clearly situational and our research in that regard is ongoing.

\subsection{Centrality measures in semantic networks}\label{app:ch1:centrality}

Recall that outgoing ties in $G^C$ give weight to the aspects of meaning that a word transmits to other words. Simultaneously, the word receives a characterization from another group of words via its incoming ties. Thus, the flow of semantic identity through the pipes of the substitution network (\cite{podolnyNetworksPipesPrisms2001}) defines three related properties. First, across all paths of the network, which flows of semantic identity ultimately accrue to a given word? Second, which words are axial in this transmission of identity? Finally, which words are especially determinant of the meaning of other words that convey a very specific meaning?

We highlight the canonical Katz-Bonacich centrality (\cite{bonacich1987power}) in what follows, but note that PageRank centrality (\cite{page1999pagerank}), while offering a similar interpretation, is generally more stable for directed and weighted networks. With slight abuse of notation, we denote the adjacency matrix of a substitution network as as $G^C$. The $[i,j]$th element of this matrix is the substitution tie $g_{i,j}(C)$ - the weighted path from $i$ to $j$. Next, consider the squared adjacency matrix $\left(G^C\right)^2=G^C G^C$. In this matrix, $[i,j]$th element is $\sum_{k} g_{i,k}(C)g_{k,j}(C)$, that is, the sum of weighed paths from $i$ to $j$ of length two. We can understand this value as the aspects of meaning that $i$ transmits to $j$ across any third node $k$. Generally, the $[i,j]$th element of $\left(G^C\right)^d$ gives the meaning $i$ imbues on $j$ through paths of length $d$.
Consider the \textit{backing-out} matrix
$$
B^C=\sum_{d=1}^{\infty} \delta^{d} \left(G^C\right)^d=\left(1- \delta G^C \right)^{-1}-1
$$
where $\delta$ is an attenuation term.\footnote{It is chosen to be less in magnitude than the reciprocal of the largest eigenvalue of $G^C$, allowing the infinite sum to converge.}
The $[i,j]$ element of $B^C$ measures the meaning that $i$ transmits to $j$, both directly and indirectly. We again conceive $B^C$ as a network, now giving the dyadic relations $b_{\tau,\mu}(C)$ between words $\mu$ and $\tau$ in terms of semantic identity that they transmit and receive throughout the semantic network. 

The Katz-Bonacich centrality of focal words $\tau$ is given by

$$\left[B^C \cdot 1\right]_{\tau}=\sum_{\mu} b_{\tau,\mu}(C)$$

The centrality is high for words that contribute large portions of meaning to words that themselves contribute strongly to the semantic identity of other words.

Finally \cite{bonacich1987power} proposes a modification of the above centrality measure, setting the attenuation term $\delta$ to a negative value. In this case, the word $\tau$ has a high \textit{power centrality}, if it transmits meaning to words that themselves are less axial in determining the meaning of other terms. If a word has a high power centrality, then it is especially determinant of the semantic identity of words that are not adequate substitutions for other words. $\tau$ would draw its semantic prominence from determining the meaning of specific, rather than general terms.

Structural features of meaning can be traced across contexts. For example, if the centrality of a word $\tau$ increases over time, $\tau$ can replace more words that themselves imbue significant aspects of meaning. Such a change is not necessarily caused by $\tau$ being used more often, although it is likely that a more central term sees more usage in the corpus. Instead, an increase in centrality is caused by $\tau$ being used in a larger set of sentences and contexts, especially those that include language applicable to convey matters of importance. However, an increase in centrality needs not be associated with the notion of generality in terms of meaning. Rather, centrality is a measure of prominence that depends on the subject matter discussed in the text. If the corpus becomes more specialized on a given subject matter, a word $\tau$ may become more central in the language because it conveys crucial elements of this prominent subject.
To distinguish prominence from generality, we next turn to a measure of brokerage across distinct meanings.

\subsubsection{Brokerage across meanings}

Consider an open triad structure in the semantic network. A focal term $\tau$ is a plausible substitution for two other words, $\mu$ and $\rho$, that can not suitably substitute for each other in the context $C$. In this constellation, $\tau$ unites elements of two distinct aspects of meaning. We may then ask whether $\tau$ holds such a structural position in more instances, and across a larger set of semantic associations.

Following substitution ties through the network $G^C$ indicates a shift in meaning from the sense of one word to the sense of the other. The Betweenness centrality (\cite{newmanMeasureBetweennessCentrality2005}) of a word $\tau$ measures how many shortest paths of meaning in $G^C$ pass through the  semantic identity of the focal word. A high Betweenness centrality indicates that $\tau$ brokers between otherwise distinct semantic identities. For example, if $\tau$ is the word \textit{manager}, and the context $C$ are sentences about groups of people, then $\tau$ might broker between roles associated with sports teams and roles associated with companies. On the one hand, a brokerage position would indicate $\tau$'s applicability in both areas. On the other hand, we can conceive brokering words like $\tau$ as those semantic properties that unite either area. In the example, both sports teams and companies share the need to manage a group of individuals.

If, for instance, the Betweenness centrality of a word increases over time, then this word is increasingly part of a wider array of disconnected conversations. If a word increasingly brokers across more aspects of meaning, its use has become more general. However, since the meaning of a word is also defined by its applicability, such generality may come at the cost of precision. The more distinct aspects of semantics a word connects, the less it conveys a specific meaning.


\subsection{Computational Implementation}

Our approach can be divided into three steps. 

First, we train the language model on the subset of the corpus of interest. We employ BERT, as implemented in the PyTorch-Transformers package (Wolf et.al. 2019), with minor modifications.  While our method would provide the best results when trained from scratch, we make concessions for the size of our corpus and available compute power by fine-tuning each division from the same BERT-base pre-trained model. Given that interpretability, not downstream performance, is the goal of the present approach, we use a modified dictionary without word-piece divisions. One BERT model is trained separately for each division of the corpus - in this case, one year of Harvard Business Review publications.

Second, we predict each word in each sentence, extracting the probability distributions and calculating a multigraph network on the fly. To do so, we repeatedly feed the same sentence into the language model, each time masking one word. That is, our inference tasks corresponds exactly to the probability distributions we use in our method. We extract ties from distributions for words that are not stop words. Given the MLM head, the distributions have full support. For that reason and for each occurrence, we capture a certain amount of the probability mass of the distribution and add ties and metadata to a graph database as discussed in more detail below. In this paper, we save $90$ percent of probability mass as ties.

For our analysis of networks, free parameters relate to the precision and sparsity of generated networks. Similarity networks are generally dense, and cut-offs improve the discriminatory properties of clustering algorithms. We can cut ties at several places. First, when querying a semantic network from our graph database, discarding low valued ties improves performance and reduces memory load. We find that this is only necessary for contextual ties, as they include queries across several positions in the sequence and tend to result in dense (often near complete) networks. We found a cutoff value of $0.1$ to induce sparsity without affecting any significant ties.
Second, a sparsity operation on the conditioned network may improve clustering performance. Here, we propose to retain a certain percentage of the total mass of outgoing ties. Even high values like $99$ percent are sufficient to discard low-valued ties that merely introduce noise in the clustering algorithm. 
However, to demonstrate that results do not significantly depend on this sparsification, results in the present paper are presented without further modifications before clustering.

\subsubsection{Technical implementation of graphs}

Our method requires us to retain a graph with a large amount of ties and a limited amount of nodes. Depending on cutoff levels and bounds on the maximum degree, the graph can grow rapidly. For that reason, the method is not an efficient approach to represent the downstream model for upstream tasks: While the graph could be embedded further, the value of doing so is questionable. Instead, it should be clear that our method is best suited to analyze either the deep learning model or the corpus with the maximal retention of relational information. It excels at this task, specifically because graphs, as opposed to embeddings, represent the entirety of relational information.

As detailed above, we do not consider links with a very small weight. Besides exponentially increasing the time of computation, small links represent noise and do not change results. After this step, we re-normalize the distributions to retain our interpretation of edge weights as probabilities. Our method can include predicted probability of the ground-truth, in which case our graphs have self-loops, or it can discard it and re-normalize across the remaining substitutions. Note that the identity of the occurring term can be recovered by the reverse tie direction.

While conditioned graphs are usually small enough to fit into memory,  the greatest flexibility can be achieved by retaining all ties and condition on demand. Thus, we employ Neo4j as graph database to save the entire multi-graph for all context and corpus divisions.
Our schema defines both tokens and edges of $G$ and $E$ as nodes in the graph database, which allows efficient sub-setting with appropriate indices.

We developed a Python interface that allows unified training, processing and querying of the networks, and implement all proposed measures in it. This framework is due to be released as a Python package, allowing replication of our methods.

\subsubsection{Model fit and modifications to saturate probability distributions}\label{app:ch1:saturation_expansion}

We seek to maximize the fit of BERT for two reasons. First, the validity of our approach hinges on the performance of the underlying language model and the initial versions of BERT were found to be undertrained (\cite{liuRoBERTaRobustlyOptimized2019}). Second, our model needs to represent a given corpus of text rather than perform out-of-sample predictions. Thus, the better the language model performs in-sample relative to out-of-sample, the more its predictions are likely to capture the idiosyncrasies of the focal text. In that sense, we strive to overfit our model, given the computational resources at hand.

Since BERT is trained to minimize cross-entropy loss, an overfit model could potentially lead to output distributions where the ground truth element approaches a probability of one. While a softmax head guarantees that the output distributions are never fully degenerate, the weights of appropriate substitutes may approach the noise floor of the distribution. 

Our experiments show that the substitute distributions do not collapse and, in part, this is due to the randomized substitutions used in the training procedure of BERT. In addition, we conjecture that the continuity in the representation space and the attention operation ultimately lead to representations that do not allow for zero entropy in the output probability distribution.

However, the present article does not seek to provide a conclusive answer to the above issues. Instead, we note that probability distributions can always be saturated if the problem arises. First, the softmax function takes a parameter (called temperature or exploration) that accomplishes such a saturation (\cite{puranamModellingBoundedRationality2015,heDeterminingOptimalTemperature2018,zhangHeatedUpSoftmaxEmbedding2018}). Second, if the dependency of entropy on training intensity becomes an issue, we propose to calculate  cross-entropy loss without considering the top $k$ predicted classes, where $k$ is drawn randomly, forcing the model to keep distributions saturated.

\newpage
\printbibliography







\end{document}